\documentclass[11pt]{article} 

\usepackage{times}
\usepackage{jmlr2e}
\usepackage{macros}
\usepackage{nicefrac}

\graphicspath{{figures/}}

\title{Efficient Change-Point Detection  \\ for Tackling Piecewise-Stationary Bandits \vspace{0.2cm}}

\author{\name{Lilian Besson}  \\
  \addr{ENS Rennes, IRISA, Inria Rennes, France}\\
  \email{lilian.besson@ens-rennes.fr}
\AND
\name{Emilie Kaufmann}\\
  \addr{Univ. Lille, CNRS, Inria,  Centrale Lille, UMR 9189 - CRIStAL, F-59000 Lille, France} \\
  \email{emilie.kaufmann@univ-lille.fr}
\AND
\name{Odalric-Ambrym Maillard}\\
\addr{Univ. Lille, Inria, CNRS, Centrale Lille, UMR 9189 - CRIStAL, F-59000 Lille, France} \\
\email{odalric.maillard@inria.fr}
\AND
\name{Julien Seznec}\\
\addr{Inria and \emph{Lelivrescolaire.fr} Éditions} \\
\email{julien.seznec@inria.fr}
}

\editor{}

\begin{document}

\maketitle

\begin{abstract}We introduce \GLRklUCB, a novel algorithm for the piecewise \iid{} non-stationary bandit problem with bounded rewards. 
    This algorithm combines an efficient bandit algorithm, \klUCB, with an efficient, \emph{parameter-free}, changepoint detector, the Bernoulli Generalized Likelihood Ratio Test, for which we provide new theoretical guarantees of independent interest. Unlike previous non-stationary bandit algorithms using a change-point detector, \GLRklUCB{} does not need to be calibrated based on prior knowledge on the arms' means. We prove that this algorithm can attain a $\cO(\sqrt{TA\Upsilon_T\log(T)})$ regret in $T$ rounds on some ``easy'' instances, where $A$ is the number of arms and $\Upsilon_T$ the number of change-points, \emph{without prior knowledge of} $\Upsilon_T$. In contrast with recently proposed algorithms that are agnostic to $\Upsilon_T$, we perform a numerical study showing that \GLRklUCB{} is also very efficient in practice, beyond easy instances. 
\end{abstract}

\begin{keywords}
    Multi-Armed Bandits; Change Point Detection; Non-Stationary Bandits. 
\end{keywords}

\section{Introduction}
\label{sec:Introduction}

Multi-Armed Bandit (MAB) problems form a well-studied class of sequential decision making problems, in which an agent repeatedly chooses an action $A_t \in\{1,\dots,A\}$ or ``arm'' 
among a set of $A$ arms \citep{Robbins52,LattimoreBanditAlgorithmsBook}. 
In the most standard version of the stochastic bandit model, each arm $a$ is associated with an \iid{} sequence of rewards $(X_{a,t})$ that follow some distribution of mean $\mu_a$. Upon selecting arm $A_t$, the agent receives the reward $X_{A_t,t}$. 
Her goal is to design a sequential arm selection strategy that maximizes the expected sum of these rewards, or, equivalently, that minimizes \emph{regret}, defined as the difference between the total sum of rewards of an oracle strategy always selecting the arm with largest mean and that of her strategy. 

Stochastic bandits were historically introduced as a simple model for clinical trials, where arms correspond to some treatments with unknown efficacy \citep{Thompson33}. More recently, MAB models have been proved useful for other applications, such as cognitive radio, where arms can model the vacancy of radio channels, or parameters of a dynamically configurable radio hardware \citep{Maghsudi16, Bonnefoi17, KerkoucheAlami18}. Another application is the design of recommender systems, where arms model the popularity of different items (\eg, news recommendation, \cite{Li10contextual}).
In all these applications, the assumption that the arms distributions \emph{do not evolve over time} is often violated: patients adapt to medical treatments, new devices can enter or leave the radio network, hence impacting the availability of radio channels, and the popularity of items is subject to trends. This aroused interest in how to take \emph{non-stationary} aspects into account within a multi-armed bandit model.

As a possible way to cope with non-stationarity, the \emph{piecewise stationary MAB} was introduced by \cite{Kocsis06}.
In this model, the (random) reward of arm $a$ at round $t$ has some mean $\mu_a(t)$ and the regret is measured with respect to the \emph{current} best arm $a_t^\star = \arg\max_{a} \mu_a(t)$. It is furthermore assumed that there are relatively few \emph{breakpoints} between which the $\mu_a(t)$ remain constant for all arms $a$.  
Despite many approaches already proposed for minimizing regret under this model (see Section~\ref{sec:BanditSetting}), research on this topic has been very active in the last years, notably in two different directions. The first is the design of a good combination of a bandit algorithm and a changepoint detector (CPD) supported by regret guarantees and enjoying \emph{good empirical performance} \citep{LiuLeeShroff17, CaoZhenKvetonXie18}. These algorithms share with many others the downside of having to know the number of breakpoints $\Upsilon_T$ to guarantee state-of-the-art regret. The second direction proposes algorithms that achieve \emph{optimal regret without the knowledge of $\Upsilon_T$} \citep{Auer19NonStat, Luo19Context}, but without an emphasis on actual practical performance (yet). 

In this paper, we propose the first algorithm based on a change-point detector that is very efficient in practice and \emph{does not require the knowledge of $\Upsilon_T$} to provably achieve optimal regret, at least on some ``easy'' instances, with few breakpoints of large enough magnitude. An interesting feature of our algorithm compared to other CPD-based algorithms is that it 
\emph{does not require any prior knowledge on the arms means}. Like \CUSUM{} \citep{LiuLeeShroff17} and \MUCB{} \citep{CaoZhenKvetonXie18}, our algorithm relies on combining a standard bandit algorithm with a changepoint detector. For the bandit component, we propose the use of the \klUCB{} \citep{KLUCBJournal} which is known to 
outperform {\UCB} \citep{Auer02} used in previous works. For the changepoint detector, we suggest using the Bernoulli Generalized Likelihood Ratio Test (GLRT), for which we provide new non-asymptotic properties that are of independent interest. This choice is particularly appealing because unlike the changepoint detectors used in previous works, the Bernoulli GLRT \emph{does not require a lower bound on the minimal amount of change to detect}, which leads to a bandit algorithm which is agnostic to the arms' means. In contrast, both \CUSUM{} and \MUCB{} 
require the knowledge of the smallest magnitude of a change in the arm's mean. 

In this work we jointly investigate two versions of \GLRklUCB{}, one using \emph{global restarts} (resetting the history of \emph{all} arms once a changepoint is detected on one of them) and one using \emph{local restarts} (resetting the history of an arm each time a changepoint is detected on that arm). We prove that \GLRklUCB{} based on global restart achieves a $\cO(\sqrt{TA\Upsilon_T \ln(T)} /(\Delta^{\text{change}})^2)$ regret where $\Delta^{\text{change}}$ is the smallest magnitude of a breakpoint. If all breakpoints have a large magnitude, this $\cO(\sqrt{TA\Upsilon_T \ln(T)})$ regret is matching the lower bound of \cite{Seznec20RestlessRotting} up to a $\sqrt{\ln(T)}$ factor. Following a similar analysis, we prove slightly weaker results for the version based on local restart. Numerical simulations in Section~\ref{sec:NumericalExperiments} reveal that these two versions are both competitive in practice with state-of-the-art algorithms.

To summarize, our contributions are the following: (1) A non-asymptotic analysis of the Bernoulli-GLR changepoint detector. (2) A new bandit algorithm for the piecewise stationary setting based on this test that needs no prior knowledge on the number of change-points and no information on the arms means to attain near-optimal regret. (3) An extensive numerical study illustrating the good performance of two versions of \GLRklUCB{} compared to other algorithms with state-of-the-art regret. 

\paragraph{Outline} The paper is structured as follows. We introduce the model and review related works in Section~\ref{sec:BanditSetting}. In Section~\ref{sec:ChangePointDetector}, we present some properties of the Bernoulli-GLR changepoint detector.
We introduce the two variants of \GLRklUCB{} in Section~\ref{sec:GLRklUCB_Algorithm}. In Section~\ref{sec:Analysis} we present regret upper bounds for \GLRklUCB{} for Global Restart and sketch our regret analysis. Numerical experiments are presented in Section~\ref{sec:NumericalExperiments}.

\section{Setup and Related Work}
\label{sec:BanditSetting}

A \emph{piecewise stationary bandit model} is characterized by a stream of (random) rewards $(X_{a,t})_{t\in\N^\star}$  associated to each arm $a \in \{1,\dots,A\}$. We assume that the rewards are bounded in a known range, and without loss of generality we assume that $X_{a,t} \in [0,1]$.
 We denote by $\mu_{a}(t) :=  \bE[X_{a,t}]$ the mean reward of arm $a$ at round $t$. At each round $t$, a decision maker has to select an arm $A_t\in\{1,\dots,A\}$, based on past observation and receives the corresponding reward $r(t) = X_{A_t,t}$. At time $t$, we denote by $a_t^\star$ an arm with maximal expected reward, \ie, $\mu_{a_t^\star}(t) = \max_a \mu_a(t)$, called an optimal arm. 
 
A policy $\pi$ chooses the next arm to play based on the sequence of past plays and obtained rewards.
The performance of $\pi$ is measured by its (dynamic) \emph{regret}, the difference between the expected reward obtained by an oracle policy playing an optimal arm $a^\star_t$ at time $t$, and that of the policy $\pi$:
\[
    R_T^{\pi} = \E\left[\sum_{t=1}^T \left(\mu_{a^\star_t}(t) - \mu_{A_t}(t)\right)\right].
\]
In the piecewise \emph{i.i.d.} model, we furthermore assume that there is a (relatively small) number of \emph{breakpoints}, denoted by $\Upsilon_T:=\sum_{t=1}^{T-1} \ind\left(\exists a\in\{1,\dots,A\}: \mu_t(a) \neq \mu_{t+1}(a)\right)$.
We define the $k$-th breakpoint by $\tau^{(k)} = \inf\{t > \tau^{(k-1)} : \exists a : \mu_a(t) \neq \mu_{a}(t+1)\}$ with $\tau^{(0)}=1$. Hence for $t\in[\tau^{(k)} + 1,\tau^{(k+1)}]$, the rewards $(X_{a,t})$ associated to all arms are \iid, with mean denoted by $\mu^{(k)}_a$. The \emph{magnitude} of a breakpoint $k$ is defined as $\Delta^{c,(k)} =: \underset{a =1,\dots,A}{\max} \left|\mu_a^{(k)} - \mu_a^{(k-1)}\right|$ and we let $\Delta^{\text{change}} =: \underset{k = 1 ,\dots, \Upsilon_T}{\min} \Delta^{c,(k)}$.

Note than when a breakpoint occurs, we do not assume that all the arms means  change, but that \emph{there exists} an arm which experiences a \emph{changepoint}, i.e. whose mean satisfies $\mu_a(t)\neq \mu_{a}(t+1)$. Depending on the application, many scenarios can be meaningful: changes occurring for all arms simultaneously (due to some exogenous event), or only a few arms that experience a changepoint in each breakpoint. Letting $C_T$ denote the total number of changepoints before horizon $T$, we have $C_T \in \{\Upsilon_T, \dots, A\Upsilon_T\}$.

\subsection{An Adversarial View on Non-Stationary Bandits} 

A natural way to cope with non-stationary is to model the decision making problem as an \emph{adversarial bandit problem} \citep{Auer02NonStochastic}, under which the rewards are arbitrarily generated.
For adversarial environments, the most studied performance measure is the pseudo-regret, which compares the accumulated reward of a given strategy with that of the best fixed-arm policy. However in some changing environments 
it is more natural to measure regret against the best \emph{sequence of actions}. \cite{Auer02NonStochastic} propose the Exp3.S algorithm, that achieves a regret of $\cO(\sqrt{A\Upsilon_T T \ln(T)})$ against the best sequence of actions with $\Upsilon_T-1$ switches. This regret rate matches the corresponding lower bound. Exp3.S is simple to implement and run with time and space complexity $\cO(A)$ but requires the knowledge of $T$ and $\Upsilon_T$ to reach near-minimax optimal regret rate.  

When the piecewise i.i.d. assumption holds (with $\Upsilon_T$ stationary part), the best sequence of actions with $\Upsilon_T-1$ switches corresponds to the optimal oracle policy. The minimax optimal rate against piecewise i.i.d. rewards sequences is also $\cO(\sqrt{A\Upsilon_T T})$. It is similar to the fixed-arm case where the adversarial pseudo-regret rate and the minimax stochastic rate are the same ($\cO(\sqrt{AT})$, \cite{audibert2010minimax}). However, in the fixed-arm setup, the stochastic stationary assumption allows a \emph{problem-dependent analysis}: some algorithms (e.g. UCB or Thomson Sampling) suffer $\cO(\nicefrac{\log{T}}{\Delta_i})$ regret on each arm $i$ with a reward gap of $\Delta_i$ compared to the best arm. When $\Delta_i$ is large enough,  this problem-dependent guarantee is much better than the $\cO(\sqrt{T})$ minimax rate. Unfortunately, in the piecewise i.i.d. setup, \cite{Garivier11UCBDiscount} show that any algorithm whose regret is $R_T(\bm\mu)$ on a stationary bandit instance $\bm\mu$ is such that there exists a piecewise stationary instance $\bm\mu'$ with at most two breakpoints such that $R_T(\bm\mu') \geq c T/R_T(\bm\mu)$, for some absolute constant $c$. In particular, this implies that an algorithm that attains $\cO(\sqrt{T})$ regret for \emph{any} piecewise stationary bandit model has no hope to reach $\cO(\log(T))$ regret on easy instances. 
The intuition behind this result is that if an algorithm achieves very low regret on a specific problem then it has to pull suboptimal arms very scarcely. By doing so, it is unable to perform well on a similar problem where the identified suboptimal arms' surreptitiously increase to become optimal. Therefore, it is important to pull every arm often enough (e.g. every $\cO(\sqrt{\nicefrac{AT}{\Upsilon_T}})$ rounds) even when one is clearly underperforming. 

Nevertheless, the piecewise i.i.d. bandit problem remained actively studied since the seminal paper of \cite{Auer02NonStochastic}. The outcome of this line of work is threefold. First, designing strategies leveraging tools from the stochastic MAB can greatly improve the empirical performance compared to adversarial algorithms like Exp3.S. Second, we would like to build strategies that are near-optimal without the knowledge of $\Upsilon_T$\footnote{In the adversarial setup, $\Upsilon_T$ appears in the definition of the pseudo-regret, hence it is quite natural that the learner knows this parameter. In the piecewise i.i.d. setup, the regret is against the optimal oracle policy which is defined independently of $\Upsilon_T$.} (unlike Exp3.S). Third, it is possible to further restrain the setup to make the problem-dependent analysis possible by forbidding the aforementioned surreptitious increase of one arm. For instance, \cite{OdalricSubo19} consider the ``global change" setup in which all the arms change significantly when a breakpoint occurs. \cite{Seznec20RestlessRotting} consider the rotting setup where the arms cannot increase. In both cases, the authors proved a logarithmic problem-dependent upper bound on the regret of their algorithms. 

In this paper, we bring theoretical and empirical contributions to the two first points. We also discuss a possible adaptation of \GLRklUCB{} that may recover logarithmic regret in the easier setups of \cite{OdalricSubo19, Seznec20RestlessRotting}.
\subsection{Algorithms Exploiting the Stochastic Assumption} \label{subsec:RelatedStochastic}
The piecewise stationary bandit model was first studied by \cite{Kocsis06,YuMannor09,Garivier11UCBDiscount}. It is also known as \emph{switching} \citep{MellorShapiro13} or \emph{abruptly changing stationary} \citep{WeiSrivastava18} environment. 
Most approaches exploiting the stochastic assumption combine a bandit algorithm with a mechanism to \emph{forget old rewards}.
We make a distinction between \emph{passively adaptive strategies}, which use a fixed forgetting mechanism, and \emph{actively adaptive strategies}, for which this mechanism is also data-dependent.

\paragraph{Passively Adaptive Strategies} A simple mechanism to forget the past consists in either discounting rewards (multiplying past reward by $\gamma^n$ where $n$ is the time elapsed since that reward was collected, for a discount factor $\gamma\in(0,1)$), or using a sliding window (only the rewards gathered in the $\tau$ last rounds are taken into account, for a window size $\tau$).
Those strategies are passively adaptive as the discount factor or the window size are \emph{fixed}, and can be tuned as a function of $T$ and $\Upsilon_T$ to achieve a certain regret bound.
Discounted UCB (D-UCB) was proposed by \cite{Kocsis06} and analyzed by \cite{Garivier11UCBDiscount}, who prove a $\cO(A\sqrt{\Upsilon_T T }\ln(T))$ regret bound, if $\gamma = 1 - \sqrt{\Upsilon_T/T}/4$.
The same authors proposed the Sliding-Window UCB (SW-UCB) and prove a $\cO(A\sqrt{\Upsilon_T T \ln(T)})$ regret bound, if $\tau = 2 \sqrt{T\ln(T)/\Upsilon_T}$. More recently, \cite{RajKalyani17} proposed the Discounted Thompson Sampling (DTS) algorithm, which performs well on the reported experiments with $\gamma=0.75$, but no theoretical guarantees are given for this particular tuning. The RExp3 algorithm \citep{Besbes14stochastic} is another passively adaptive strategy that is based on (non-adaptive) restarts of the Exp3 algorithm \citep{Auer02NonStochastic}. RExp3 is analyzed in terms of a different measure of interest, the total variation budget $V_T$ which satisfies $\Delta^{\text{change}} \Upsilon_T \leq V_T \leq \Upsilon_T$. RExp3 is proved to have a $\cO((A\log{A})^{1/3} V_T^{1/3} T^{2/3})$ regret, which translates to a sub-optimal rate in our setting. 

\paragraph{Actively Adaptive Strategies}

The first \emph{actively adaptive} strategy is Windowed-Mean Shift \citep{YuMannor09}, which combines any bandit policy with a change point detector which performs \emph{adaptive restarts} of the bandit algorithm. However, this approach does not apply to our setting as it takes into account side observations. Another line of research on actively adaptive algorithms uses a Bayesian point of view,
where the process of change point occurrences is modeled and tracked using Bayesian updates.
A Bayesian Change-Point Detection (CPD) algorithm is combined with Thompson Sampling by \cite{MellorShapiro13}, and more recently in the Memory Bandit algorithm of \cite{Alami17}. Since none of these algorithms have theoretical guarantees and they are designed for a different setup, we do not include them in our experiments. 
%

Our closest competitors rather use frequentist CPD algorithms combined with a bandit algorithm. The first algorithm of this flavor, Adapt-EVE algorithm \citep{Hartland06} uses a Page-Hinkley test and the \UCB{} policy, but no theoretical guarantees are given. Exp3.R \citep{Allesiardo15, Allesiardo17} combines a CPD with Exp3, and the history of all arms are reset as soon as a sub-optimal arm is detected to become optimal and it achieves a $\cO(\Upsilon_T A\sqrt{T \ln(T)})$ regret (without the knowledge of $\Upsilon_T$). More recently, \CUSUMUCB{} \citep{LiuLeeShroff17} and Monitored UCB (\MUCB, \cite{CaoZhenKvetonXie18}) have achieved $\cO(\sqrt{\Upsilon_T A T \ln(T)})$ regret, when $\Upsilon_T$ is known. 


\CUSUMUCB{} is based on a variant of a two-sided \CUSUM{} test, that uses the first $M$ samples from one arm to compute an initial average, and then detects whether a drift of size larger than $\varepsilon$ occurred from this value by checking whether a random walk based on the remaining observations crosses a threshold $h$. 
It requires the tuning of three parameters, $M$, $\varepsilon$ and $h$. \CUSUMUCB{} performs \emph{local restarts} using this test, to reset the history of \emph{one arm} for which the test detects a change.
\MUCB{} uses a simpler test, based on the $w$ most recent observations from an arm: a change is detected if the absolute difference between the empirical means of the first and second halves of those $w$ observations exceeds a threshold $h$. It requires the tuning of two parameters, $w$, and $h$. \MUCB{} performs \emph{global restarts} using this test, to reset the history of \emph{all arms} whenever the test detects a change on one of them. 

On a stationary batch, a UCB index algorithm tends to pull each arm at a logarithmic rate asymptotically. According to the aforementioned \cite{Garivier11UCBDiscount}'s lower bound, it is not enough to shield against increases of the suboptimal arms' values. Thus, CPD-based algorithms usually rely on additional \emph{forced exploration}: each arm is pulled regularly either according to a constant probability of uniform exploration \citep{LiuLeeShroff17} or according to a deterministic scheme \citep{CaoZhenKvetonXie18}. To avoid linear regret, the total budget dedicated to this forced exploration is tuned with the knowledge of $T$ and $\Upsilon_T$  (e.g. $\cO\left(\sqrt{A\Upsilon_TT}\right)$). \cite{OdalricSubo19} suggest canceling the forced exploration when all the arms change at the same rounds. Indeed, in that case, we can aim to detect the changes on any arms' sequences and then restart all the arms' indexes. Similarly, \cite{Seznec20RestlessRotting} do not use forced exploration and study the Rotting Adaptive Window UCB (RAW-UCB) - a UCB index policy with an adaptive window designed for non-increasing sequences of rewards. Both these algorithms can get logarithmic regret on some problem instances and, therefore, cannot be minimax optimal for the general piecewise i.i.d bandit problem, which is our focus in this paper.  



\subsection{Knowledge of the Number of Breakpoints} 

All algorithms mentioned above for the general piecewise stationary bandit problem require some tuning that should depend on $\Upsilon_T$ to attain state-of-the-art $\cO(\sqrt{\Upsilon_T AT\ln(T)})$ regret. Two algorithms achieving this regret \emph{without the knowledge of }$\Upsilon_T$ were recently proposed: Ada-ILTCB$^+$\citep{Luo19Context} and AdSwitch \citep{Auer19NonStat}, that also rely on detecting non-stationarities \citep{AuerLuo19}.
While the former is tailored for the more general adversarial and contextual setting, the latter is specifically proposed for the piecewise i.i.d. model. 

AdSwitch is an elimination strategy based on confidence interval (like Improved UCB \citep{AuerOrtner10}) with global restarts when a change-point is detected on one arm. AdSwitch performs an \emph{adaptive forced exploration} scheme on the eliminated arms which adds two main components to the aforementioned uniform random exploration. First,  AdSwitch uses a counter $l$ (initialized at $1$) for the number of detected changes by the CPD subroutine as a proxy for $\Upsilon_T$ to tune the random exploration probability on each eliminated arms at $\cO\left(\sqrt{\nicefrac{l}{KT}}\right)$. Second,  AdSwitch also selects at random a change size $\Delta$ on a geometric grid, with a probability proportional to $\Delta$. The arm is then pulled $\cO\left(\nicefrac{1}{\Delta^2}\right)$ consecutive pulls to check if there is a change of size $\Delta$. The consecutive sampling is particularly helpful theoretically to analyze the algorithm when the CPD misses some breakpoints.

However, unlike in our work, the underlying changepoint detectors used in AdSwitch have not been optimized for efficiency or tractability\footnote{Indeed, at each time step $t$, the test employed by AdSwitch requires $\Theta(A t^3)$ operations, resulting in a very expensive $\Theta(A T^4)$ time complexity when compared to $\Theta(A T)$ for simple algorithms like UCB and $\Theta(A T^2)$ for other adaptive approaches based on scan statistics like \GLRklUCB{}.}. Neither \citep{Luo19Context} nor \citep{Auer19NonStat} report simulation to assess the empirical or numerical efficiency of their algorithms. In this paper, we include (a tweaked, tractable version of) AdSwitch in our experiments for short horizons.   

An alternative idea to adapt to $\Upsilon_T$ is the ``Bandit over Bandit'' approach of \cite{Cheung19NonStatLinear}, which uses an exponential weights algorithm for expert aggregation on top of several copies of Sliding-Windows UCB with different (fixed) window size. Yet this approach does not yield optimal regret.

\section{\!The Bernoulli GLR Change Point Detector}
\label{sec:ChangePointDetector}

%

Sequential changepoint detection has been extensively studied in the statistical community (see, \eg, \cite{Basseville93,gupta2000parametric,wu2007inference})). In this article, we are interested in detecting changes on the mean of a probability distribution with bounded support. Assume that  we collect independent samples $X_1,X_2,\ldots$ all from some distribution supported in $[0,1]$. We want to discriminate between two possible scenarios: all the samples come from distributions that have a common mean $\mu_0$, or there exists a \emph{changepoint} $\tau \in\N^\star$ such that $X_1,\ldots,X_\tau$ have some mean $\mu_0$ and $X_{\tau +1},X_{\tau+2},\ldots$ have a different mean $\mu_1 \neq \mu_0$.
A sequential changepoint detector is a stopping time $\hat\tau$ with respect to the filtration $\cF_t = \sigma(X_1,\dots,X_t)$ such that $(\hat \tau < \infty)$ means that we reject the hypothesis
$\cH_0 : \left(\exists \mu_0 \in [0,1]: \forall i\in\N, \bE[X_i] = \mu_0\right)$.

Generalized Likelihood Ratio tests have been used for a very long time (see, e.g. \cite{Wilks38}) and were for instance studied for changepoint detection by \cite{siegmund1995using}. Exploiting the fact that bounded distributions are $(1/4)$-sub-Gaussian (\ie, have a moment generating function dominated by that of a Gaussian with the same mean and variance $1/4$), the (Gaussian) GLRT, recently studied in depth by \cite{Maillard2018GLR}, can be used for our problem. We propose instead to exploit the fact that bounded distributions are also dominated by Bernoulli distributions.
We call a \emph{sub-Bernoulli distribution} any distribution $\nu$ that satisfies
$\ln \bE_{X\sim\nu}\left[e^{\lambda X}\right] \leq \phi_{\mu}(\lambda)$ with $\mu=\bE_{X\sim \nu}[X]$ and $\phi_{\mu}(\lambda) = \ln(1-\mu + \mu e^\lambda)$ is the log moment generating function of a Bernoulli distribution with mean $\mu$. Lemma 1 of \cite{KLUCBJournal} establishes that any bounded distribution supported in $[0,1]$ is a sub-Bernoulli distribution.

\subsection{Presentation of the test}

If the samples $(X_t)$ were all drawn from a Bernoulli distribution, our changepoint detection problem would reduce to a parametric sequential test of
$\cH_0 : (\exists \mu_0: \forall i\in\N, X_i \ \overset{\text{i.i.d.}}{\sim} \ \cB(\mu_0))$
against the alternative
$\cH_1 : (\exists \mu_0 \neq \mu_1, \tau \in \N^\star: \ \  X_1, \ldots, X_\tau \ \overset{\text{i.i.d.}}{\sim} \ \cB(\mu_0) \ \ \text{and} \ \ X_{\tau+1}, X_{\tau+2}, \ldots \ \overset{\text{i.i.d.}}{\sim} \ \cB(\mu_1))$.
The (log)-Generalized Likelihood Ratio statistic for this test is defined by
\[\GLR(n) :=\ln\tfrac{\sup\limits_{\mu_0,\mu_1,\tau < t}\ell(X_1, \ldots, X_n ; \mu_0,\mu_1,\tau)}{\sup\limits_{\mu_0}\ell(X_1, \ldots, X_n ; \mu_0)},\]
where $\ell(X_1, \ldots, X_n ; \mu_0)$ and $\ell(X_1, \ldots, X_n ; \mu_0,\mu_1,\tau)$ denote the likelihood of the first $n$ observations under a model in $\cH_0$ and $\cH_1$.
High values of this statistic tend to indicate rejection of $\cH_0$.
Using the form of the likelihood for Bernoulli distribution, this statistic can be written with the binary relative entropy $\kl$,
\begin{equation}\label{BernoulliDivergence}
    \kl(x,y) := x \ln\left(\tfrac{x}{y}\right) + (1-x)\ln\left(\tfrac{1-x}{1-y}\right).
\end{equation}
Indeed, one can show that $\GLR(n) = \sup_{s \in [1,n]} Z_{s,n}$ where $Z_{s,n} = s \times \kl\left(\hat{\mu}_{1:s},\hat{\mu}_{1:n}\right) + (n-s) \times \kl\left(\hat{\mu}_{s+1:n},\hat{\mu}_{1:n}\right)$ 
and for $k \leq k'$, $\hat{\mu}_{k:k'}$ denotes the average of the observations collected between the instants $k$ and $k'$. This motivates the definition of the Bernoulli GLR change point detector.

\begin{definition}\label{def:GLRDef}
    The Bernoulli GLR change point detector with threshold function $\beta(n,\delta)$ is
    \vspace*{-5pt}
    \begin{equation}\label{def:GLR}
        \hat\tau_{\delta} := \inf \Bigg\{ n \in \N^\star : \sup_{s \in [1,n]} \Big[s \times \kl\left(\hat{\mu}_{1:s},\hat{\mu}_{1:n}\right) + (n-s) \times \kl\left(\hat{\mu}_{s+1:n},\hat{\mu}_{1:n}\right)\Big] \geq \beta(n,\delta)\Bigg\}.
    \end{equation}
\end{definition}

Asymptotic properties of the GLR for changepoint detection have been studied by \cite{LaiXing10} for Bernoulli distributions and more generally for one-parameter exponential families, for which the GLR test is defined as in~\eqref{def:GLR} but with $\kl(x,y)$ replaced by the Kullback-Leibler divergence between two elements in that exponential family that have mean $x$ and $y$. For example, the Gaussian GLR studied by \cite{Maillard2018GLR} corresponds to \eqref{def:GLR} with $\kl(x,y) = 2(x-y)^2$ when the variance is set to $\sigma^2=1/4$, and non-asymptotic properties of this test are given for any $(1/4)$-sub-Gaussian samples. 


In the next section, we provide new non-asymptotic results about the Bernoulli GLR test under the assumption that the samples $(X_t)$ come from a sub-Bernoulli distribution, which holds for any distribution supported in $[0,1]$.
Note that Pinsker's inequality gives that $\kl(x,y) \geq 2(x-y)^2$, hence the Bernoulli GLR may stop earlier that the Gaussian GLR based on the quadratic divergence $2(x-y)^2$.

\paragraph{GLR versus confidence-based CPD}
An alternative to the GLR also based on scan statistics, used by \cite{OdalricSubo19} 
consists in building individual confidence intervals for the mean in each segment, of the form 
\[\left[\hat{\mu}_{1:s} \pm \sqrt{\frac{\tilde \beta(s,\delta)}{2s}}\right] \ \text{and} \ \left[\hat{\mu}_{s+1:n} \pm \sqrt{\frac{\tilde \beta(n-s,\delta)}{2(n-s)}}\right]\]
and report that there is a change point if there exists $s$ such that these confidence interval are disjoint, i.e. {\small
\begin{eqnarray*}\hat\tau'_\delta &=& \inf\left\{n \in \N^\star : \exists s\in [1,n] , \Big|\hat\mu_{1:s} - \hat\mu_{s+1:n}\Big| > \sqrt{\frac{\tilde{\beta}(s,\delta)}{2s}} + \sqrt{\frac{\tilde{\beta}(n-s,\delta)}{2(n-s)}}\right\}.
\end{eqnarray*}}
\hspace{-0.2cm} By measuring distances with the appropriate KL divergence function, the Bernoulli GLR test better exploits the geometry of (sub-)Bernoulli distributions.  

\subsection{Properties of the Bernoulli GLR}\label{subsec:PropGLR}

In Lemma~\ref{lem:FalseAlarm} below, we propose a choice of the threshold function $\beta(n,\delta)$ under which the probability that there exists a \emph{false alarm} under \iid{} data is small. To define $\beta$, we introduce the function $\cT$, originally introduced by \cite{KK18Martingales}, 
\begin{equation}\label{def:function_T}
    \cT(x) ~:=~ 2 \tilde h\left(\frac{h^{-1}(1+x) + \ln(2\zeta(2))}{2}\right)
\end{equation}
where for $u \ge 1$ we define $h(u) = u - \ln(u)$ and its inverse $h^{-1}(u)$.
And for any $x \ge 0$, $\tilde h(x) = e^{1/h^{-1}(x)} h^{-1}(x)$ if $x \ge h^{-1}(1/\ln (3/2))$ and $\tilde{h}(x) = (3/2) (x-\ln(\ln (3/2)))$ otherwise. The function $\cT$ is easy to compute numerically.
Its use for the construction of concentration inequalities that are uniform in time is detailed in~\cite{KK18Martingales}, where tight upper bounds on the function $\cT$ are also given:  $\cT (x) \simeq x + 4 \ln\big(1 + x + \sqrt{2x}\big)$ for $x\geq 5$ and $\cT(x) \sim x$ when $x$ is large. The proof of Lemma~\ref{lem:FalseAlarm} 
is given in Appendix~\ref{proof:FalseAlarm}.

\begin{lemma}\label{lem:FalseAlarm}
    Assume that there exists $\mu_0 \in [0,1]$ such that $\bE[X_t] = \mu_0$ and that $X_i \in [0,1]$ for all $i$. Then the Bernoulli GLR test satisfies $\bP_{\mu_0}(\hat\tau_\delta < \infty) \leq \delta$ with the threshold function
    \begin{equation}\label{def:beta}
        \beta(n,\delta)= 2\cT\left(\frac{\ln(3n\sqrt{n}/\delta)}{2}\right) + 6\ln(1+\ln(n)).
    \end{equation}
\end{lemma}

Another key feature of a changepoint detector is its \emph{detection delay} under a model in which a change from $\mu_0$ to $\mu_1$ occurs at time $\tau$. We already observed that from Pinsker's inequality, the Bernoulli GLR stops earlier than a Gaussian GLR. Hence, one can leverage some techniques from \cite{Maillard2018GLR} to upper bound the detection delay of the Bernoulli GLR. Letting $\Delta = |\mu_0 - \mu_1|$, one can essentially establish that for $\tau$ larger than $(1/\Delta^2)\ln(1/\delta)$ (\ie, enough samples before the change), the delay can be of the same magnitude (\ie, enough samples after the change). In our bandit analysis to follow, the detection delay will be crucially used to control the probability of the ``good event'' that all the changepoints are detected within a reasonable delay (Lemma~\ref{lem:GoodEventGlobal} and \ref{lem:GoodEvent}).

\subsection{Practical considerations}\label{sub:PracticalConsiderations}

Lemma~\ref{lem:FalseAlarm} provides the first control of false alarm for the Bernoulli GLR employed for bounded distributions. However, the threshold \eqref{def:beta} is not fully explicit as the function $\cT(x)$ can only be computed numerically.
Note that for sub-Gaussian distributions, results from \cite{Maillard2018GLR} show that the smaller and more explicit threshold
$\beta(n,\delta) = \left(1 + \frac{1}{n}\right)\ln\left(\frac{3n\sqrt{n}}{\delta}\right)$,
can be used to prove an upper bound of $\delta$ for the false alarm probability of the GLR, with quadratic divergence $\kl(x,y)=2(x-y)^2$.
For the Bernoulli GLR, numerical simulations suggest that the threshold \eqref{def:beta} is a bit conservative, and in practice we recommend to keep only the leading term and use $\beta(n,\delta) = \ln(n\sqrt{n}/\delta)$.


Also note that, as any test based on scan-statistics, the GLR can be costly to implement: at every time step, it considers all previous time steps as a possible position for a changepoint. Thus, in practice the following adaptation may be interesting, based on down-sampling the possible time steps:
\begin{equation}\label{def:GLRTricks}
    \tilde \tau_{\delta} = \inf \left\{ n \in \cN : \sup_{s \in \cS_n} Z_{s,n} \geq \beta(n,\delta)\right\},
\end{equation}
for any strict subsets $\cN\subseteq \N$ and $\cS_n \subset \{1,\dots,n\}$. Following the proof of Lemma~\ref{lem:FalseAlarm}, we can easily see that this variant enjoys the exact same false-alarm control. However, the detection delay may be slightly increased. Our experiments reveal that the price in terms of regret of the speed-up is negligible. 




\section{The \GLRklUCB{} Algorithm}
\label{sec:GLRklUCB_Algorithm}

\GLRklUCB{} (Algorithm~\ref{algo:GLRklUCB}) combines the \klUCB{} algorithm \citep{KLUCBJournal}, known to be optimal for Bernoulli bandits, with the Bernoulli GLR changepoint detector introduced in Section~\ref{sec:ChangePointDetector}. It also needs a third ingredient: some extra exploration to ensure each arm is sampled enough and changes can also be detected on arms currently under-sampled by \klUCB. This forced exploration is parameterized by a sequence $\alpha=(\alpha_k)_{k\in\N}$ of exploration frequencies $\alpha_k\in(0,1)$. \GLRklUCB{} can be used in any bandit model with bounded rewards, and is expected to be very efficient for Bernoulli distributions. 

    \begin{small}
    \begin{algorithm}[h]
        \begin{small}
        \KwIn{$(\alpha_k)_{k\in \N^\star}$ (sequence of exploration probabilities), \\ 
        $\delta \in (0,1)$  (maximum error probability for the test)\;}
        \KwIn{\emph{Option}: \textbf{\textcolor{blue}{Local}} or \textbf{\textcolor{red}{Global}} restart\;}
        \textbf{Initialization: } $\forall a \in \{1,\dots,A\}$, $\tau_a \leftarrow 0$ (last restart) and $n_a \leftarrow 0$ (number of selections since last restart) \\
        \hspace{2.2cm} $k \gets 1$ (number of episodes)\\
        \For{$t=1,2,\ldots, T$}{
            \uIf{
                $t \mod \left\lfloor \frac{A}{\alpha_k}\right\rfloor \in \{1,\dots,A\}$
            }{
                $A_t \leftarrow t \mod \left\lfloor \frac{A}{\alpha_k}\right\rfloor$
            }
            \Else{
                $A_t \leftarrow \arg\max\limits_{a\in \{1,\dots,A\}} \mbox{UCB}_a(t)$ as defined in \eqref{def:UCB}
            }
            \mbox{Play arm } $A_t$ \mbox{and receive the reward } $X_{A_t,t}$ : $n_{A_t} \leftarrow n_{A_t} + 1; Y_{A_t, n_{A_t}} \leftarrow X_{A_t,t}$ \\
            \uIf{
                $\GLR_\delta(Y_{A_t,1}, \ldots, Y_{A_t, n_{A_t}}) = \mathrm{True} \ $
            }
            {
                \uIf{\emph{Local} restart}{
                \textcolor{blue}{$\tau_{A_t} \leftarrow t$ and $n_{A_t} \leftarrow 0$ and $k \gets k+1$
                    }
                    }
                \Else{
                \textcolor{red}{$\forall a\in \{1,\dots,A\}, \tau_a \leftarrow t$ and $n_a \leftarrow 0 $ and $k \gets k+1$
                    }
                }
            }
        }
        \caption{\GLRklUCB{} (\textbf{Local} or \textbf{Global} restarts)\label{algo:GLRklUCB}}
        \end{small}
    \end{algorithm}
    \end{small}

The \GLRklUCB{} algorithm can be viewed as a \klUCB{} algorithm allowing for some \emph{restarts} on the different arms. A restart happens when the Bernoulli GLR changepoint detector detects a change on the arm that has been played (line $9$).
To be fully specific, $\GLR_\delta(Y_1,\dots,Y_n) = \mathrm{True}$ if and only if the GLR statistic associated to those $n$ samples,
\begin{small}
    \[\sup_{1 \leq s \leq n} \left[s \times \kl (\hat{Y}_{1:s},\hat{Y}_{1:n}) + (n-s) \times \kl (\hat{Y}_{s+1:n},\hat{Y}_{1:n})\right], \]
\end{small}%
is larger than the threshold $\beta(n,\delta)$ defined in \eqref{def:beta}, or $\beta(n,\delta) = \ln(n^{3/2}/\delta)$, as recommended in Section~\ref{sub:PracticalConsiderations}. Each restart (on any arm) triggers a new episode and we denote by $k_t$ the number of episodes started after $t$ samples (i.e. the index of the on-going episode at time $t$).

Letting $\tau_a(t)$ denote the instant of the last restart that happened for arm $a$ before time $t$, $n_a(t) = \sum_{s=\tau_a(t)+1}^{t} \ind(A_s=a)$ the number of selections of arm $a$ and $\hat{\mu}_a(t) = (1/n_a(t)) \sum_{s=\tau_a(t)+1}^{t}X_{a,s} \ind(A_s=a)$ the empirical mean (if $n_a(t)\neq 0$), the index used by the algorithm is defined as\begin{small}
\begin{equation}\label{def:UCB}
    \UCB_a(t)\! := \!\max \bigl\{ q: n_a(t) \times \kl\left(\hat{\mu}_a(t),q\right) \leq f(t \!-\! \tau_a(t)) \bigr\}.
\end{equation}\end{small}
\hspace{-0.2cm} Algorithm~\ref{algo:GLRklUCB} presents two variants of \GLRklUCB{}, one using \emph{local restarts} (line $11$), and one using \emph{global restarts} (line $13$).
Under local restarts, in the general case the times $\tau_a(t)$ are not equal for all arms, hence the index policy associated to \eqref{def:UCB} is \emph{not} a standard \UCB{} algorithm, as each index uses a \emph{different exploration rate}.
One can highlight that in the \CUSUMUCB{} algorithm, which is the only existing algorithm based on local restarts, the \UCB{} index are defined differently\footnote{This choice is currently not fully supported by theory, as we found mistakes in the analysis of \CUSUMUCB{}: Hoeffding's inequality is wrongly used with a \emph{random} number of observations and a \emph{random} threshold to obtain Eq. $(31)$-$(32)$.}:
$f(t-\tau_a(t))$ is replaced by $f(n_t)$ with $n_t = \sum_{a=1}^A n_a(t)$.

The forced exploration scheme used in \GLRklUCB{} (lines $3$-$5$) generalizes the deterministic exploration scheme proposed for \MUCB{} by \citep{CaoZhenKvetonXie18}, whereas \CUSUMUCB{} performs randomized exploration.
A consequence of this forced exploration is given in Proposition~\ref{prop:EnoughSamples} (proved in Appendix~\ref{proof:EnoughSamples}).

\begin{proposition}\label{prop:EnoughSamples}
Let $s,t$ be two time instants between  two consecutive restarts on arm $a$ (i.e. $\tau_a(t) < s < t$). Then it holds that $n_a(t) - n_a(s) \geq \left\lfloor \frac{\alpha_{k_t}}{A} (t-s) \right\rfloor$, with $k_t$  the number of episodes before round $t$.
\end{proposition}

\section{Regret Analysis}\label{sec:Analysis}

In this section, we prove regret bounds for \GLRklUCB{} using Global Restart. Our results for \GLRklUCB{} with Local Restart are a bit weaker and are deferred to Appendix~\ref{app:Local}.  

\subsection{Regret Upper Bounds}

Recall that $\tau^{(k)}$ denotes the position of the $k$-th breakpoint and let $\mu_a^{(k)}$ be the mean of arm $a$ on the segment $\{ \tau^{(k-1)}+1, \dots, \tau^{(k)} \}$. We also introduce $k^\star = \arg\max_a \mu_a^{(k)} $, the sub-optimality gap $\Delta_a^{(k)} = \mu_{k^\star}^{(k)} - \mu_{a}^{(k)}$ and the recall that the magnitude of breakpoint $k$ is $\Delta^{c,(k)} = \underset{{a=1,\dots,A}}{\max} \ |\mu_a^{(k)} - \mu_a^{(k-1)}| >0$.

We first introduce an assumption, which is easy to interpret and standard in non-stationary bandits. It requires that the distance between two consecutive breakpoints is large enough: how large depends on the magnitude of the largest change that happens at those two breakpoints.

\begin{assumption}\label{ass:LongPeriodsGlobal}
    Define the delay $d^{(k)} = d^{(k)}(\alpha,\delta)$ as
    \[d^{(k)}(\alpha,\delta) = \left\lceil \tfrac{4A}{\alpha_k\left(\Delta^{c,(k)}\right)^2}\beta\left(\tfrac{3}{2}(\tau^{(k)}-\tau^{(k-1)}),\delta\right) + \tfrac{A}{\alpha_k} \right\rceil,\]
    we assume that $\forall k\leq \Upsilon_T, \tau^{(k)} - \tau^{(k-1)} \geq 2(d^{(k)}\vee d^{(k-1)})$.
\end{assumption}

Under Assumption~\ref{ass:LongPeriodsGlobal}, we provide in Theorem~\ref{thm:mainRegretBoundGlobal} a finite time problem-dependent regret upper bound. It features the parameters $\alpha$ and $\delta$,
the gaps $\Delta_a^{(k)}$ and KL-divergence terms $\kl(\mu_{a}^{(k)},\mu_{k^\star}^{(k)})$ expressing the hardness of the stationary MAB problem between two breakpoints,
and the detection delays $d^{(k)}(\alpha,\delta)$, which feature the gap $\Delta^{c,(k)}$ and express the hardness of the detection of each breakpoint.

%
%

\begin{theorem}
    \label{thm:mainRegretBoundGlobal}
    For $\alpha=(\alpha_1,\alpha_2,\dots)$ an increasing exploration sequence and $\delta$ for which Assumption~\ref{ass:LongPeriodsGlobal} is satisfied, the regret of \GLRklUCB{} with parameters $\alpha$ and $\delta$ based on \textbf{Global} Restart satisfies{\small
\begin{align*}&R_T \leq  (A+1)\Upsilon_T \delta T  + A\delta T + \alpha_{\Upsilon_T+1} T  \\
& + \sum_{k=0}^{\Upsilon_T}\sum_{a : \Delta_a^{(k)}>0} \!\!\!\min \left\{\Delta_a^{(k)}\left(\tau^{(k+1)}-\tau^{(k)}\right) ; \Delta_a^{(k)} \left[d^{(k)}(\alpha,\delta) + \frac{\ln\left(\tau^{(k+1)}-\tau^{(k)}\right)}{\kl\left(\mu_a^{(k)},\mu_{k^\star}^{(k)}\right)} \right] + \bigO{\sqrt{\ln\left(\tau^{(k+1)}-\tau^{(k)}\right)}}\right\}.\end{align*}}
\end{theorem}

We express below the scaling of this regret bound when the exploration sequence $\alpha$ and the parameter $\delta$ are carefully tuned using the knowledge of the horizon $T$, but \emph{without the knowledge of the number of breakpoints} $\Upsilon_T$. We express this scaling as a function of the smallest value of a sub-optimality gap on one of the stationary segments and the gap of the hardest breakpoint to detect, respectively defined as  $\Delta^{\text{opt}} := \underset{k = 1,\dots, \Upsilon_T}{\min}\underset{\{a : \Delta_a^{(k)}>0\}}{\min} \ \Delta_a^{(k)}$, and $\Delta^{\text{change}} := \underset{k = 1,\dots, \Upsilon_T}{\min} \Delta^{c,(k)}= \underset{k = 1,\dots, \Upsilon_T}{\min} \ \underset{a=1,\dots,A}{\max} \ |\mu_a^{(k)} - \mu_a^{(k-1)}|$.

\begin{corollary}\label{cor:Global}
For any $\alpha_0 \in \R^+$ and $\gamma \in (1/2,1]$, choosing \[\alpha_k = \alpha_0\sqrt{\frac{kA\ln(T)}{T}} \ \ \ \text{and} \ \ \ \delta = \frac{1}{T^\gamma},\]
on problem instances satisfying the corresponding Assumption~\ref{ass:LongPeriodsGlobal}, the regret of \GLRklUCB{} satisfies 
\[R_T  = \bigO{(1+\gamma)\frac{\sqrt{\Upsilon_T AT \ln(T)}}{\left(\Delta^{\emph{change}}\right)^2} + \frac{(A-1)}{\Delta^{\emph{opt}}} \Upsilon_T\ln(T)},\]
and $R_T = \bigO{\frac{\sqrt{\Upsilon_T AT \ln(T)}}{\left(\Delta^{\emph{change}}\right)^2}}.$
\end{corollary}

%

If $\Delta^{\text{change}}$ is viewed as a constant, our regret upper bound is matching the lower bound of \cite{Seznec20RestlessRotting} up to a $\sqrt{\log(T)}$ factor. Hence we propose a tuning of \GLRklUCB{} which attains near-optimal regret without the knowledge of the number of breakpoints, on ``easy'' problems such that two consecutive breakpoints are separated by more than $\sqrt{TA\ln(T)}/(\Delta^{\text{change}})^2$ time steps. As shown is Section~\ref{sec:NumericalExperiments}, this doesn't prevent \GLRklUCB{} from performing well on more realistic instances, which was
similarly observed by \citet{CaoZhenKvetonXie18} for \MUCB{}.
The dependency in $\left(\Delta^{\text{change}}\right)^{-2}$ is also present in the regret bound for other algorithms combining UCB-style algorithms and changepoint detectors \cite{LiuLeeShroff17,CaoZhenKvetonXie18}. It may come from a limitation of the current analysis of such algorithms, which require every breakpoint to be detected. 

Compared to other algorithms based on stationary bandit strategies combined with change-point detectors, \GLRklUCB{} is the only one that doesn't require a tuning based on $\Upsilon_T$ to attain the best possible regret. Indeed, it uses an increasing exploration sequence instead of a constant sequence, which allows the trick \eqref{TheTrickIsHere} in the proof of Corollary~\ref{cor:Global}. If $\Upsilon_T$ is known, observe that one can also run \GLRklUCB{} with the constant exploration sequence $\alpha_k = \alpha = \alpha_0\sqrt{{kA\ln(T)}/{T}}$ and obtain the same regret as in Corollary~\ref{cor:Global}. We tried the two alternatives in our experiments, and got similar performances. Hence, the use of an exploration sequence that is agnostic to $\Upsilon_T$ does not hinder the practical performance of \GLRklUCB{}. 

Finally, there exist algorithms which attain near-optimal regret without the knowledge of $\Upsilon_T$ and with no  $\left(\Delta^{\text{change}}\right)^{-2}$ multiplicative factor \citep{Auer19NonStat,Luo19Context}. However, these algorithms are very conservative in order to make their analysis possible. For instance, AdSwitch is an elimination policy, which is often a poor choice in practice for regret minimization. In Section~\ref{sec:NumericalExperiments}, we indeed show that \GLRklUCB{} greatly outperforms AdSwitch.

\subsection{Proof of Corollary~\ref{cor:Global}} With the choice $\delta = T^{-\gamma}$ and $\alpha_k = \alpha_0\sqrt{{k\ln(T)A}/{T}}$, Theorem~\ref{thm:mainRegretBoundGlobal} yields the following upper bound on the regret of \GLRklUCB{}: 
{\small
\begin{align}&(A+1) \Upsilon_T T^{1-\gamma} + A T^{-\gamma} + \alpha_0\sqrt{(\Upsilon_T+1)  A T\ln(T)} + \sum_{k=1}^{\Upsilon_T}  \frac{4A}{\alpha_k\left(\Delta^{c,(k)}\right)^2} \beta(\tfrac{3}{2}T,T^{-\gamma}) \nonumber \\ &+ \sum_{k=1}^{\Upsilon_T}\sum_{a : \mu_a^{(k)}\neq \mu_{k^\star}^{(k)}} \frac{\left( \mu_{k^\star}^{(k)}-\mu_a^{(k)}\right)}{\kl\left(\mu_a^{(k)},\mu_{k^\star}^{(k)}\right)}\ln(T) + \bigO{\sqrt{\ln(T)}}\;.\label{DoSomethingDifferent}
\end{align}}

\hspace{-0.2cm} For $\gamma > 1/2$, the leading term in this expression is 
\[\alpha_0\sqrt{(\Upsilon_T+1)  A T\ln(T)} + \sum_{k=1}^{\Upsilon_T}  \frac{4A}{\alpha_k\left(\Delta^{c,(k)}\right)^2} \beta(\tfrac{3}{2}T,T^{-\gamma}) + \sum_{k=1}^{\Upsilon_T}\sum_{a : \mu_a^{(k)}\neq \mu_{k^\star}^{(k)}} \frac{\left( \mu_{k^\star}^{(k)}-\mu_a^{(k)}\right)}{\kl\left(\mu_a^{(k)},\mu_{k^\star}^{(k)}\right)}\ln(T)\;.\]
Using that $\beta(n,\delta) \leq C \ln(n/\delta)$ for some absolute constant $C$ together with the fact that
\begin{equation}\sum_{k=1}^{\Upsilon_T}\frac{1}{\alpha_k} = \frac{1}{\alpha_0}\sqrt{\frac{T}{A\ln(T)}}\sum_{k=1}^{\Upsilon_T} \frac{1}{\sqrt{k}} \leq \frac{1}{\alpha_0}\sqrt{\frac{\Upsilon_T T}{A\ln(T)}}\label{TheTrickIsHere}\end{equation}
yields the following control on the expected regret {\small
\begin{align*}R_T & = \bigO{(1+\gamma)\frac{\sqrt{\Upsilon_T AT \ln(T)}}{\left(\min_{k=1}^{\Upsilon_T}\Delta^{c,(k)}\right)^2} + \sum_{k=1}^{\Upsilon_T}\sum_{a : \mu_a^{(k)}\neq \mu_{k^\star}^{(k)}} \frac{\left( \mu_{k^\star}^{(k)}-\mu_a^{(k)}\right)}{\kl\left(\mu_a^{(k)},\mu_{k^\star}^{(k)}\right)}\ln(T)}\;.
\end{align*}} 
\hspace{-0.2cm} The conclusion follows from Pinsker's inequality: $\kl(\mu_a^{(k)},\mu_{k^\star}^{(k)}) \geq 2\left(\Delta_a^{(k)}\right)^2$ and from lower bounding all sub-optimality gaps by $\Delta^{\text{opt}}$.

Rather than using the problem-dependent complexity of the MAB problem on each stationary segment, using Theorem~\ref{thm:mainRegretBoundGlobal} and standard techniques one can also obtain the following ``worse-case'' upper bound:
{\footnotesize\begin{align*}&(A+1) \Upsilon_T T^{1-\gamma} + AT^{-\gamma} + \alpha_0\sqrt{(\Upsilon_T+1)  A T\ln(T)} + \sum_{k=1}^{\Upsilon_T}  \frac{4A}{\alpha_k\left(\Delta^{c,(k)}\right)^2} \beta(\tfrac{3}{2}T,T^{-\gamma}) + \sum_{k=1}^{\Upsilon_T}\sqrt{A\left(\tau^{(k+1)} - \tau^{(k)}\right)\ln(T)}\,.
\end{align*}}
\hspace{-0.2cm} Using the Cauchy-Schwarz inequality, the last term in this sum is upper bounded by $\sqrt{\Upsilon_T AT\ln(T)}$. Following the same steps as before, we get a scaling in the regret that no longer depends on $\Delta^{\text{opt}}$.

\subsection{Proof of Theorem~\ref{thm:mainRegretBoundGlobal}}
\label{sec:RegretAnalysis}

We first introduce some notation for the proof. Recall that $\tau^{(k)}$ denote the $k$-th breakpoint, we add the convention that $\tau^{(\Upsilon_T+1)} = T$. We denote by $\hat\tau^{(k)}$ the $k$-th breakpoint detected by \GLRklUCB. 

Distinguishing the exploration steps and the steps in which \GLRklUCB{} uses the UCBs to select the next arm to play, one can upper bound the regret as{\small
\begin{align} R_T  \leq  \bE\left[\sum_{t=1}^T \ind{\left(t \left\lceil \frac{A}{\alpha_{k_t}} \right\rceil \in \{1,\dots,A\}\right)} +  \sum_{t=1}^T (\mu_{a_t^\star}(t) - \mu_{A_t}(t))\ind{\left(\UCB_{A_t}(t-1) \geq \UCB_{a_t^\star}(t-1)\right)}\right] \label{twotermsGlobal}
\end{align}}
\hspace{-0.2cm} We now introduce some high-probability event in which all the breakpoints are detected within a reasonable delay. With $d^{(k)}=d^{(k)}(\alpha,\delta)$ in Assumption~\ref{ass:LongPeriodsGlobal}, we define
\begin{equation}\cE_T = \cE_T(\alpha,\delta) := \left(\forall k \in \{1, \ldots, \Upsilon_T\}, \hat{\tau}^{(k)} \in \left[\tau^{(k)} + 1, \tau^{(k)} + d^{(k)}\right], \hat{\tau}^{\left(\Upsilon_T+1\right)} > T\right).\label{def:GoodEvenGlobal}\end{equation}
Note that from Assumption~\ref{ass:LongPeriodsGlobal}, as the period between two changes are long enough, if $\cE_T$ holds, then for all change $k$, one has $\tau^{(k)} \leq \hat{\tau}^{(k)} \leq \tau^{(k+1)}$ for all $k \in \{1,\dots,\Upsilon_T\}$. 
Also, when $\cE_T$ holds, \GLRklUCB{} experiences exactly $\Upsilon_T$ restarts which permits to upper bound the exploration term in \eqref{twotermsGlobal}, using the convention that $\hat \tau^{(\Upsilon_T+1)} = T$:
\begin{eqnarray*}\sum_{t=1}^T \ind{\left(t \left\lceil \frac{A}{\alpha_{k_t}} \right\rceil \in \{1,\dots,A\}\right)} &\leq& \sum_{k=0}^{\Upsilon_T} \sum_{t=\hat{\tau}^{(k)} +1}^{\hat\tau^{(k+1)}} \ind\left(t \left\lceil \frac{A}{\alpha_{k+1}} \right\rceil \in \{1,\dots,A\}\right) \\
&\leq& \sum_{k=0}^{\Upsilon_T} \alpha_{k+1}\left(\hat{\tau}^{(k+1)} - \hat{\tau}^{(k)}\right) \leq \alpha_{\Upsilon_T + 1 }\sum_{k=0}^{\Upsilon_T} \left(\hat{\tau}^{(k+1)} - \hat{\tau}^{(k)}\right) \\
& = & \alpha_{\Upsilon_T + 1 }  T.
\end{eqnarray*}
On $\cE_T$, the second term in \eqref{twotermsGlobal} can also be decomposed along the $\Upsilon_T+1$ episodes experienced by the algorithm. Recalling that $k^\star$ denotes the optimal arm for $t \in \left[\tau^{(k)} +1,\tau^{(k+1)}\right]$, one can write
{\small
\begin{align} R_T &\leq  T \bP\left(\cE_T^c\right) + \alpha_{\Upsilon_T + 1 }  T + \sum_{k=0}^{\Upsilon_T} \bE\left[\ind(\cE_T) \!\!\sum_{t={\tau}^{(k)}+1}^{\tau^{(k+1)}}\!\! \left(\mu_{k^\star}^{(k)} - \mu_{A_t}^{(k)}\right)\ind{\left(\UCB_{A_t}(t-1) \geq \UCB_{k^\star}(t-1)\right)} \right]\label{term:banditGlobal}.
\end{align}}

The conclusion follows from the two lemmas stated below, whose proofs are given in Appendix~\ref{app:Global}. The first one hinges on some elements of the analysis of the \klUCB{} algorithm proposed by \cite{KLUCBJournal} whereas the second exploits the changepoint detection mechanism. 

\begin{lemma}\label{lem:BanditGlobal} With $\Delta_a^{(k)} = \mu_{k^\star}^{(k)} - \mu_a^{(k)}$, the following upper bound holds:
\small{\[\eqref{term:banditGlobal} \leq \sum_{k=0}^{\Upsilon_T}\sum_{a : \Delta_a^{(k)}>0}\!\!\! \min \left\{\Delta_a^{(k)}\left(\tau^{(k+1)}-\tau^{(k)}\right) ; d^{(k)} + \frac{\Delta_a^{(k)}\ln\left(\tau^{(k+1)}-\tau^{(k)}\right)}{\kl\left(\mu_a^{(k)},\mu_{k^\star}^{(k)}\right)} + \bigO{\sqrt{\ln\left(\tau^{(k+1)}-\tau^{(k)}\right)}}\right\}\;.\]}
\end{lemma}

\begin{lemma}\label{lem:GoodEventGlobal}
    Under Assumption~\ref{ass:LongPeriodsGlobal}, it holds that $\bP(\cE_T^c(\alpha,\delta)) \leq \delta(A+1)\Upsilon_T + A\delta$.
\end{lemma}

The tricky part in the analysis is the proof of Lemma~\ref{lem:GoodEventGlobal}, which crucially exploits Assumption~\ref{ass:LongPeriodsGlobal}, that we briefly sketch here (with a detailed proof in Appendix~\ref{app:CrucialLemma}). 
Introducing the event $\cC^{(k)} = \left\{\forall \ell \leq k, \hat\tau^{(\ell)} \in \left[\tau^{(\ell)} + 1, \tau^{(\ell)} + d^{(\ell)} \right] \right\}$ that all the changes up to the $k$-th have been detected and using the convention $\tau^{(\Upsilon_T +1)} = T$, a union bound permits to upper bound $\bP(\cE_T^c)$ by the sum of two terms:
\begin{small}
\[
    \sum\limits_{k=1}^{\Upsilon_T+1}\! \underbrace{\bP\left(\hat\tau^{(k)} \leq \tau^{(k)}\!\! \;|\; \!\cC^{(k-1)}\!\right)}_{(a)}\! + \!\sum\limits_{k=1}^{\Upsilon_T}\! \underbrace{\bP\left(\hat\tau^{(k)} \geq \tau^{(k)} + d^{(k)} \;|\; \cC^{(k-1)}\right)}_{(b)}.
\]
\end{small}
\hspace{-0.2cm} The event in $(a)$ implies that the change point detector associated with some arm $a$ experiences a false alarm. The probability of such an event is upper bounded by Lemma~\ref{lem:FalseAlarm} for a changepoint detector run in isolation. Under the bandit algorithm, arm $a$'s change point detector is based on less than $t - \tau_a(t)$ samples, which makes a false alarm even less likely. We finally show that $(a) \leq A\delta$ (with union bound over the $A$ arms).

Term $(b)$ is related to the control of the detection delay, which is more involved under the \GLRklUCB{} adaptive sampling scheme,
when compared to a result like Theorem~6 in \cite{Maillard2018GLR} for the changepoint detector run in isolation.
More precisely, we need to leverage the forced exploration (Proposition~\ref{prop:EnoughSamples}) to be sure we have enough samples for detection. This explains why the detection delay for the $k$-th breakpoint defined in Assumption~\ref{ass:LongPeriodsGlobal} is scaled by $\alpha_k$.
Using some elementary calculus and a concentration inequality given in Lemma~\ref{lem:Chernoff2arms}, we can finally prove that $(b) \leq \delta$.

\section{Experimental Results}
\label{sec:NumericalExperiments}

In this section, we report numerical simulations performed on synthetic data to compare the performance of \GLRklUCB{} against other state-of-the-art approaches. Experiments were performed with a library written in the Julia language which is available online.\footnote{\href{https://github.com/EmilieKaufmann/PiecewiseStationaryBandits}{https://github.com/EmilieKaufmann/PiecewiseStationaryBandits}}

\paragraph{Algorithms and Parameters Tuning}
\label{sub:ParametersTuning}
We include two baselines: the \klUCB{} algorithm (not designed for the non-stationary setting) and an algorithm that we call Oracle-\klUCB, which knows the exact locations of the breakpoints, and restarts \klUCB{} for all arms at those locations. Then, we include algorithms with state-of-the-art regret for the piecewise stationary MAB presented in Section~\ref{sec:BanditSetting}. For a fair comparison, all algorithms that use  \UCB{} as a sub-routine were adapted to use \klUCB{} instead, which yields better performance\footnote{\cite{LiuLeeShroff17,CaoZhenKvetonXie18} both mention that extending their analysis to the use of \klUCB{} should not be too difficult.}. For all the algorithms, we used the tuning recommended in the corresponding paper, using in particular the knowledge of the number of breakpoints $\Upsilon_T$ and the horizon $T$ when needed. Only two algorithms do not require the knowledge of $\Upsilon_T$: \AdSwitch{} and \GLRklUCB{}.  

We experiment with Exp3.S (with theoretically optimal tuning in Corollary 8.3 of \cite{Auer02NonStochastic}),  and the two passively adaptive algorithms Discounted \klUCB{} (D-\klUCB) with discount factor $\gamma = 1 - \sqrt{\Upsilon_T/T}/4$ and Sliding-Window \klUCB{} (SW-\klUCB) with window-size $\tau = \lceil 2 \sqrt{T\ln(T)/\Upsilon_T}\rceil $. As for \emph{actively adaptive algorithms}, we experiment with $\mathrm{AdSwitch}$ \citep{Auer19NonStat} and three algorithms combining a change-point detector with $\klUCB{}$: \CUSUMklUCB{}, \MklUCB{} and \GLRklUCB{}. These three algorithms share the use of an exploration parameter that we call $\alpha$ (or an exploration sequence $\alpha_k$ for \GLRklUCB{}). \cite{LiuLeeShroff17} and \cite{CaoZhenKvetonXie18} recommend two slightly different tuning for \CUSUMklUCB{} and \MklUCB{} respectively, that both scale in $\sqrt{\Upsilon_T\log(T)/T}$. This is also the order of magnitude of $\alpha_{\Upsilon_T}$ given by Corollary~\ref{cor:Global} for \GLRklUCB{}. Hence, in order to compare algorithm that adds a similar amount of exploration, we set $\alpha = \sqrt{{\Upsilon_TA\ln(T)}/{T}}$ for all algorithms using a constant exploration probability and $\alpha_k = \sqrt{{k A\ln(T)}/{T}}$ for the exploration sequence of \GLRklUCB{}.

Regarding the parameters of the change-point detectors, we use a threshold $h = \ln(T/\Upsilon_T)$ for \CUSUMklUCB{}, as recommended by \cite{LiuLeeShroff17}, and experience with different values of $(M,\epsilon)$ that have to be tuned using some prior knowledge of the problem. 
For \MklUCB{}, we experience with different values of the windows parameter $w$ (often choosing the tuning $w=800$ that was found to be robust in the experiments of \cite{CaoZhenKvetonXie18}) and use the recommended threshold $b=\sqrt{w \ln(2 A T^2)/2}$. For the Bernoulli-GLR test, we use the threshold function $\beta(n,\delta) = \ln(n^{3/2}/\delta)$ and set $\delta = 1/\sqrt{T}$, which is the largest value licensed by Corollary~\ref{cor:Global}.

For \GLRklUCB{} and \AdSwitch{}, which are computationally more demanding due to the use of tests based on scan-statistics, we use some implementation tweaks. For \GLRklUCB{}, we use some down-sampling as discussed in Section~\ref{sub:PracticalConsiderations}, performing the test only every $\Delta t =10$ time steps and scanning every  $\Delta s = 5$ observations for a possible change-point. 
To be able to implement \AdSwitch{} up to a horizon $T=5000$, we used $\Delta t = 50$. The computational bottleneck in \AdSwitch{} is the checks on good arms that compare the empirical mean between $s$ and $t$ to that between $s_1$ and $s_2$ for all possible $s < t$ and $s_1 \leq s_2 < t$. We only test values of $s_1$, $s_2$ and $s$ satisfying $s' \mod \Delta s = 0$, for $\Delta s = 20$. This reduces the time complexity by $(\Delta s)^3$, which is a significant speed-up in practice. 
Finally, the parameter $C_1$ that governs the elimination of good arms and should be chosen large enough was set to $C_1 = 1$.

\begin{figure}[t]{
 \centering 
 \includegraphics[width=0.54\linewidth]{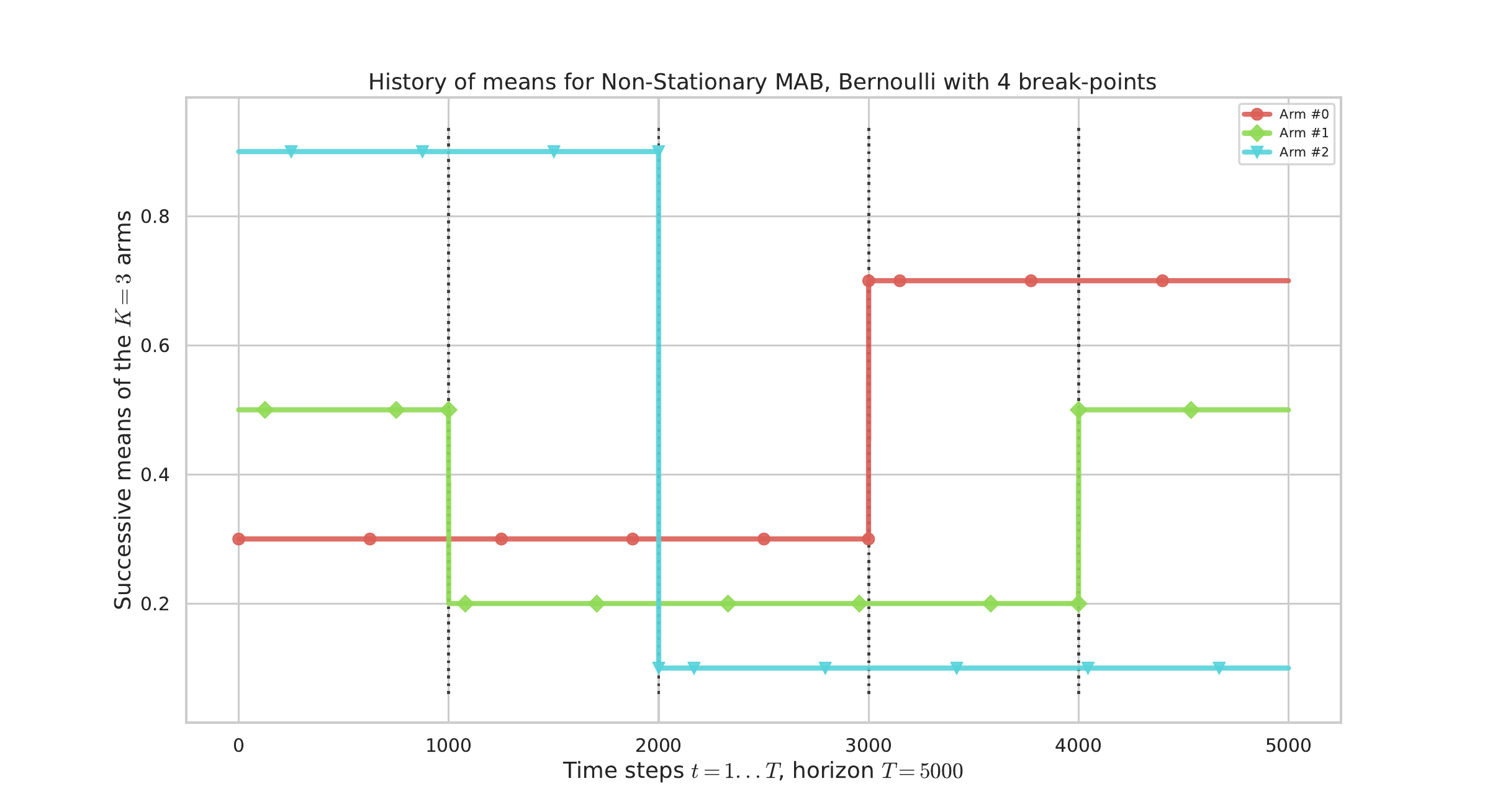}
  \includegraphics[width=0.45\linewidth]{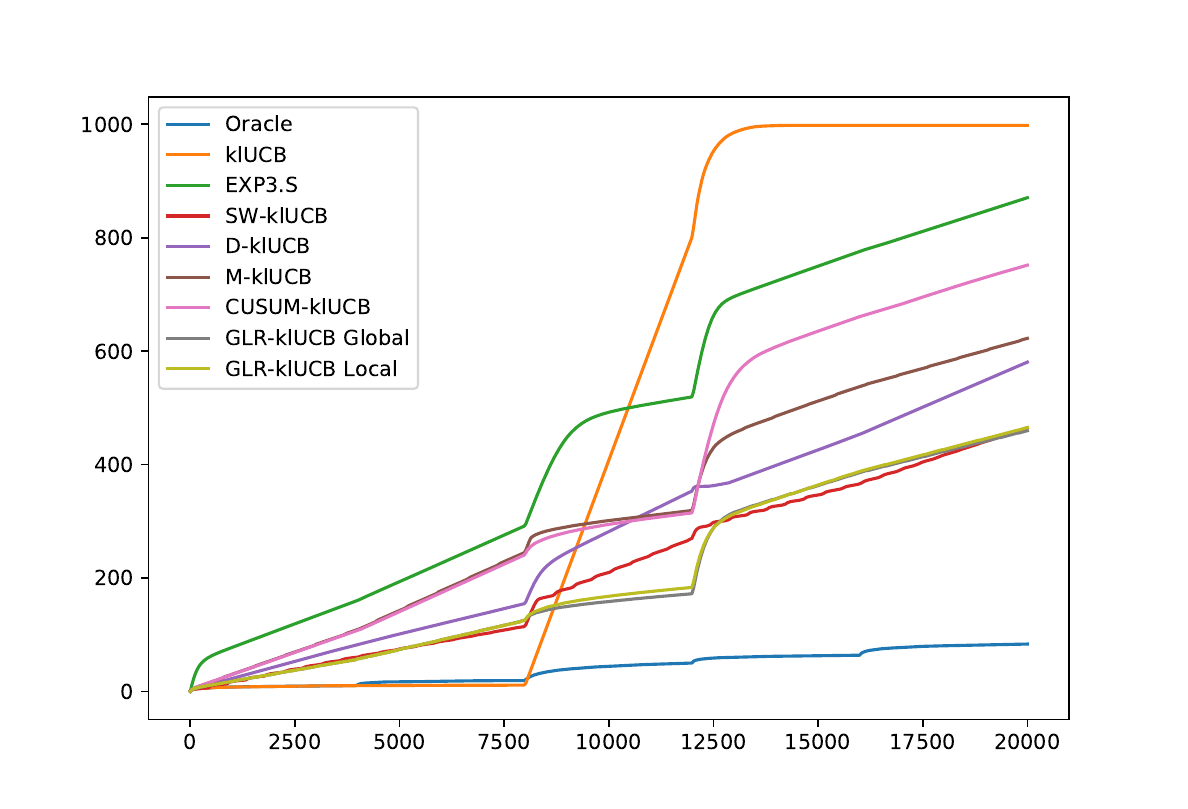}
  
  \vspace{-0.6cm}
  
\footnotesize
    \begin{center}
    \begin{tabular}{|*{9}{c|}}
    \hline
     \textbf{Algorithm} &  \multicolumn{2}{c|}{$T = 5000$} & \multicolumn{2}{c|}{$T = 10000$} & \multicolumn{2}{c|}{$T = 20000$}  & \multicolumn{2}{c|}{$T = 100000$} \\
    \hline 
      &  \multicolumn{2}{c|}{$(\alpha \simeq 0.142)$} & \multicolumn{2}{c|}{$(\alpha \simeq 0.105)$} & \multicolumn{2}{c|}{$(\alpha \simeq 0.077)$} & \multicolumn{2}{c|}{$(\alpha \simeq 0.019)$} \\
      \hline 
        Oracle-\klUCB{} &  $59 \pm 11$ &$4$ & $70 \pm 13$ & $4$ & $84 \pm 15$ & $4$ & $114 \pm 17$ & $4$ \\
      \hline  
      \klUCB{} & $270 \pm 66$ & $0$ & $515 \pm 110$ & $0$ & $998 \pm 202$ & $0$ & $4769 \pm 784$ & $0$ \\
      \hline
      EXP3.S &  $379 \pm 31$ & $0$ & $578 \pm 44$ & $0$ & $870 \pm 64$ & $0$ & $2125 \pm 142$ & $0$ \\
      SW-\klUCB{} & $\bm{186 \pm 19}$ & $0$ & $\bm{298 \pm 29}$ & $0$ & $\bm{465 \pm 37}$ & $0$ & $\bm{1269 \pm 74}$ & $0$ \\
      D-\klUCB{} & $216 \pm 14$  & $0$ & $355 \pm 19$ & $0$ & $581 \pm 26$ & $0$ & $1731 \pm 55$ & $0$ \\
      \hline
      AdSwitch & $1339 \pm 96$ & $2.2$ & $-$ & $-$ & $-$ & $-$ & $-$ & $-$ \\
     \hline
      M-\klUCB{} ($w = 200$) & $273 \pm 32$ & $1.8$ & $408 \pm 50$ & $1.9$ & $612 \pm 89$ & $1.9$ & $1848 \pm 532$ & $1.8$\\
      M-\klUCB{} ($w = 800$)  & $280 \pm 34$ & $1.2$ & $415 \pm 60$ & $1.4$ & $623 \pm 92$ & $1.8$ & $1811 \pm 343$ & $2$ \\
      CUSUM-\klUCB{} ($M \!=\! 200, \varepsilon \!=\! 0.1$) & $280 \pm 50$ & $3.2$ & $422 \pm 79$ & $4.1$ & $646 \pm 123$ & $5$ & $1776\pm 265$ & $5.8$  \\
      CUSUM-\klUCB{} ($M \!=\! 400, \varepsilon\!=\!0.05$) & $321 \pm 89$ & $4.6$ & $485 \pm 119$ & $7.1$ & $752 \pm 179$ & $11.1$ & $2018 \pm 457$ & $28.5$\\
      \hline
    \GLRklUCB{} Global ($\alpha_k$)&  $\bm{199 \pm 38}$ & $2$ & $\bm{290 \pm 59}$ & $2$ & $\bm{460 \pm 95}$ & $2.1$ & $1420 \pm 340$ & $2.8$ \\
      \GLRklUCB{} Global (constant $\alpha$) & $268 \pm 33$ & $2$ & $403 \pm 59$ & $2.2$ & $609 \pm 91$ & $2.5$ & $1804 \pm 348$ & $2$  \\
      \GLRklUCB{} Local ($\alpha_k$) & $\bm{200 \pm 39}$ & $2.1$ & $\bm{295 \pm 59}$ & $2.2$ & $\bm{465 \pm 103}$ & $2.4$ & $1403 \pm 354$ & $3$\\
      \GLRklUCB{} Local  (constant $\alpha$) & $269 \pm 32$ & $2.3$ & $399 \pm 52$ & $2.6$ & $602 \pm 82$ & $2.9$ & $1786 \pm 334$ & $3$\\
      \hline
    \end{tabular}
  \caption{\label{fig:Pb1}\small Results for \textbf{Problem 1} (displayed in the top left corner). The table show the final regret $R_T$ and the average number of restarts for several algorithms run for different horizon $T$. The top right corner displays the cumulative regret of the top 8 algorithms for $T=20000$. The regret is estimated based on $N=1000$ independent repetitions (except for $T=100000$ where $N=100$ and for the AdSwitch algorithm for which $N=50$).}
\end{center}}

\vspace{-0.4cm}

\end{figure}

\begin{figure}[t]{
 \centering 
 \includegraphics[width=0.54\linewidth]{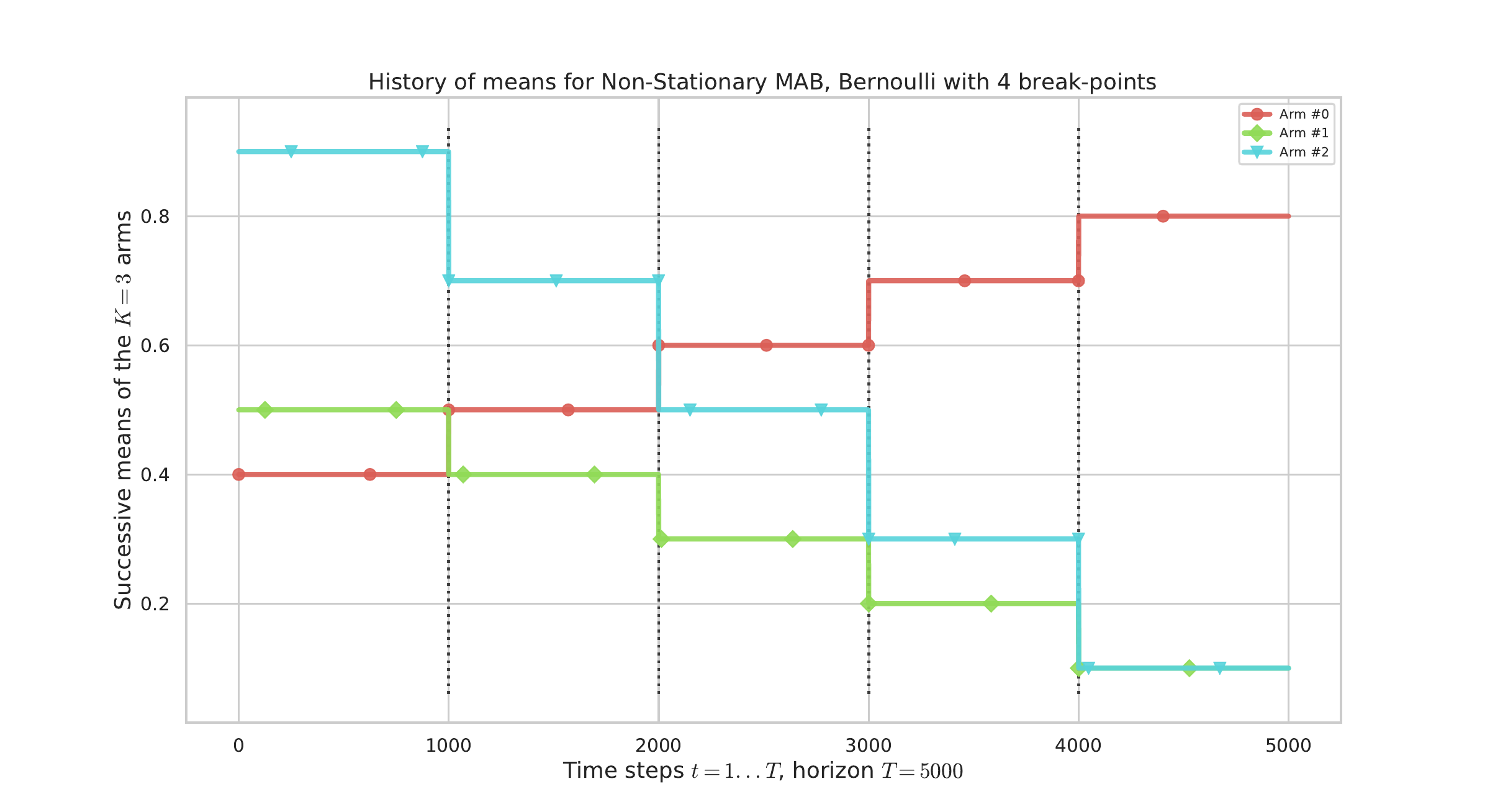}
 \includegraphics[width=0.45\linewidth]{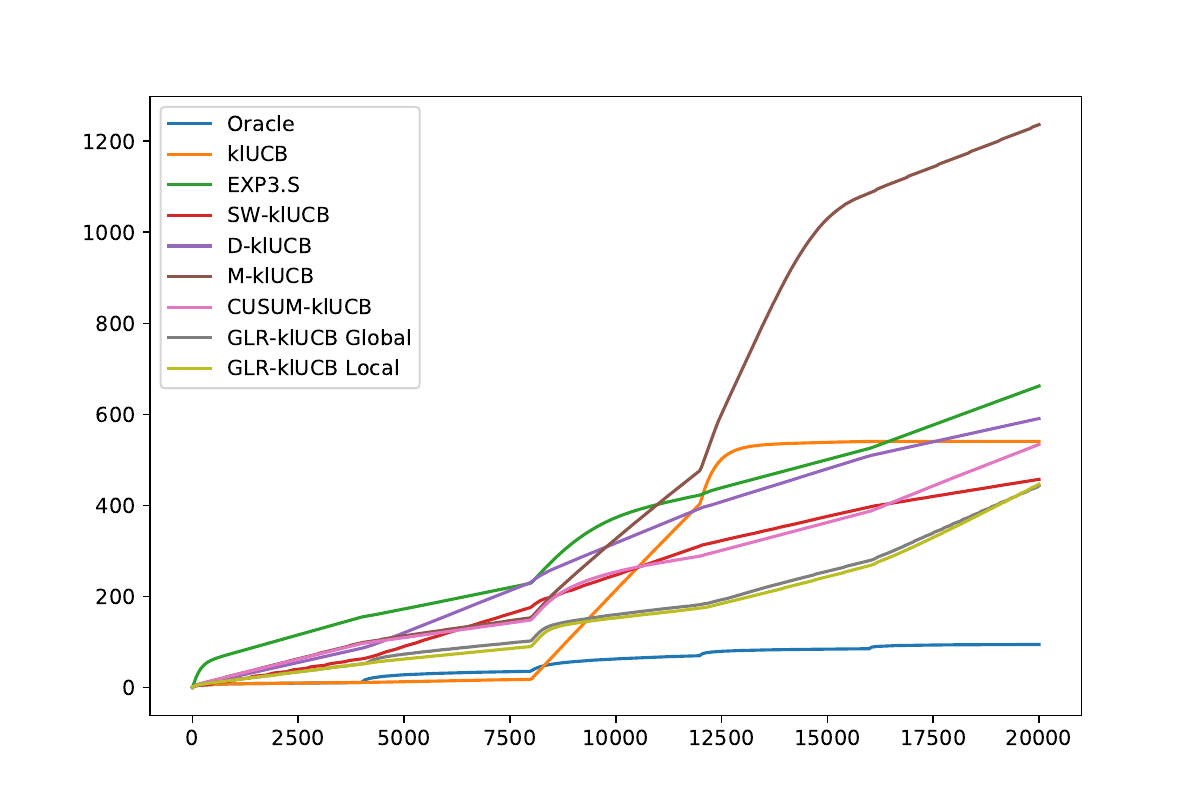}

 \vspace{-0.6cm}
 
\footnotesize
    \begin{center}
    \begin{tabular}{|*{9}{c|}}
    \hline
     \textbf{Algorithm} &  \multicolumn{2}{c|}{$T = 5000$} & \multicolumn{2}{c|}{$T = 10000$} & \multicolumn{2}{c|}{$T = 20000$}  & \multicolumn{2}{c|}{$T = 100000$} \\
    \hline 
      &  \multicolumn{2}{c|}{$(\alpha \simeq 0.142)$} & \multicolumn{2}{c|}{$(\alpha \simeq 0.105)$} & \multicolumn{2}{c|}{$(\alpha \simeq 0.077)$} & \multicolumn{2}{c|}{$(\alpha \simeq 0.019)$} \\
      \hline 
        Oracle-\klUCB{} & $65 \pm 14$ & $4$ & $80 \pm 17$ & $4$ & $95 \pm 18$ & $4$ & $127 \pm 24$ & $4$ \\
      \hline  
      \klUCB{}  & $\bm{162 \pm 52}$ & $0$ & $\bm{293 \pm 82}$ & $0$ & $541 \pm 125$ & $0$ & $2425 \pm 440$ & $0$ \\
      \hline
      EXP3.S  &  $304 \pm 30$ & $0$ & $448 \pm 42$ & $0$ & $662 \pm 56$ & $0$ & $1587 \pm 137$ & $0$ \\
      SW-\klUCB{} & $182 \pm 15$ & $0$ & $\bm{288 \pm 22}$ & $0$ & $\bm{457 \pm 29}$ & $0$ & $1258 \pm 65$ & $0$\\
      D-\klUCB{} & $218 \pm 15$ & $0$ & $361 \pm 21$ & $0$ & $591 \pm 29$ & $0$ & $1787 \pm 61$ & $0$  \\
      \hline
      AdSwitch & $1418 \pm 113$ & $1.3$   & $-$ & $-$ & $-$ & $-$ & $-$ & $-$ \\
     \hline
      M-\klUCB{} ($w = 200$) & $485 \pm 89$ & $0$ & $927 \pm 163$ & $0$ & $1813 \pm 285$ & $0$ & $9674 \pm 1082$ & $0$\\
      M-\klUCB{} ($w = 800$)  & $333 \pm 112$ & $1$ & $609 \pm 272$ & $0.8$ & $1237 \pm 617$ & $0.6$ & $7726 \pm 3331$ & $0.4$ \\
      CUSUM-\klUCB{} ($M \!=\! 200, \varepsilon \!=\! 0.1$) & $238 \pm 25$ & $6$ & $352 \pm 39$ & $7.7$ & $520 \pm 64$ & $8.6$ & $1249 \pm 191$ & $12.1$ \\
      CUSUM-\klUCB{} ($M \!=\! 400, \varepsilon\!=\!0.05$) & $238 \pm 27$ & $5.7$ & $353 \pm 51$ & $9.1$ & $534 \pm 92$ & $15.1$ & $1270 \pm 198$ & $38.5$ \\
      \hline
      \GLRklUCB{} Global ($\alpha_k$)& $209 \pm 13$ & $3$ & $\bm{303 \pm 16}$ & $3.5$ & $\bm{443 \pm 16}$ & $4$ & $\bm{989 \pm 20}$ & $4$ \\
      \GLRklUCB{} Global (constant $\alpha$) & $252 \pm 13$ & $3$ & $350 \pm 14$ & $3.5$ & $501 \pm 15$ & $4$ & $1115 \pm 17$ & $4$  \\
      \GLRklUCB{} Local ($\alpha_k$) & $206 \pm 17$ & $3.2$ & $\bm{298 \pm 23}$ & $4.1$ & $\bm{447 \pm 28}$ & $5.8$ & $\bm{1058 \pm 96} $ & $5.5$ \\
      \GLRklUCB{} Local  (constant $\alpha$) &  $249 \pm 19$ & $3.1$ & $349 \pm 26$ & $4.5$ & $506 \pm 43$ & $5.2$ & $1230 \pm 243$ & $4.9$ \\
      \hline
    \end{tabular}
    \caption{\label{fig:Pb2}\small Results for \textbf{Problem 2} (displayed in the top left corner). The table show the final regret $R_T$ and the average number of restarts for several algorithms run for different horizon $T$. The top right corner displays the cumulative regret of the best algorithms for $T=20000$. The regret is estimated based on $N=1000$ independent repetitions (except for $T=100000$ where $N=100$ and for the AdSwitch algorithm for which $N=50$).}
\end{center}}

\vspace{-0.4cm}

\end{figure}

\paragraph{Results on two simple benchmarks}
\label{sub:NumericalResults}
We design two simple piecewise stationary bandit problems with $A=3$ arms and $\Upsilon_T=4$ breakpoints. These breakpoints are evenly spaced up to the horizon, for which we investigate 4 values for each problem: $T=5000$, $T=10000$, $T=20000$ and $T=100000$. In \textbf{Problem 1}, a single arm changes in each breakpoint ($\Upsilon_T = C_T = 4$) and $\Delta^{\text{change}} = 0.3$, whereas in \textbf{Problem 2}, all arms means change at every breakpoint ($\Upsilon_T=4, C_T = 16$) and $\Delta^{\text{change}} = 0.2$. For each problem, we display the reward functions of each arm in the top left corner of Figures~\ref{fig:Pb1} and~\ref{fig:Pb2}. 

For the different values of the horizon, the reward functions are simply expanded: the size $\Delta^{\text{change}}$ remains the same and the breakpoints are still evenly spaced. Hence, it is a way to vary the difficulty of the underlying change-point detection problems. Indeed, when $T$ goes larger, the distance between two consecutive breakpoints increases, and Assumption~\ref{ass:LongPeriodsGlobal} is closer to be satisfied. On this simple problems with equally spaced breakpoints ($\tau^{(k)} - \tau^{(k-1)} = T/5$), with our choice of $\alpha_k$ and $\delta$, Assumption~\ref{ass:LongPeriodsGlobal} amounts to
\[\sqrt{T} \geq \frac{80\sqrt{A}}{\min_{k=1}^{\Upsilon_T} \sqrt{k} (\Delta^{c,(k)})^2} \sqrt{\ln \left(T\right)} + 1\]
which is only satisfied for $T$ much larger than 100000 for both Problem 1 and Problem 2. Therefore, in this experiment, we investigate the performance of \GLRklUCB{}  for difficult problems on which it does not have theoretical guarantees. 


In Figure~\ref{fig:Pb1} (respectively Figure~\ref{fig:Pb2}), we display the results for Problem~1 (respectively Problem~2). We display the regret of each algorithm as a function of the rounds for one horizon (top right corner); and we also tabulate the regret at the horizon and the number of restarts for all the algorithms and all the horizons. In Problem 1, we observe that the regret of \GLRklUCB{} with Global and Local restart is competitive with that of the SW-\klUCB{} which performs best for the different time horizons $T$. However, recall that this algorithm is tuned using the knowledge of $\Upsilon_T$, unlike \GLRklUCB{}. In Problem 2, the regret of \GLRklUCB{} is the smallest for large horizons ($T=20000,100000$) whereas for shorter horizons ($T=5000,10000$) \klUCB{} and SW-\klUCB{} have (slightly) smaller regret. Regarding other passively adaptive approaches, we see that D-\klUCB{} is competitive with (sometimes even better than) actively adaptive algorithms, whereas Exp3.S only manages to outperform \klUCB{} for large horizons. \GLRklUCB{}  largely outperforms  \AdSwitch{} for $T=5000$, which is the largest horizon for which we could implement this algorithm. We now turn our attention to CPD-based algorithms.

The tests used by \CUSUMUCB{} and \MUCB{} depend on two sets of parameters that should in principle be chosen according to some prior knowledge of the problem, and we tried for each algorithm two different tunings of these parameters. For \CUSUMUCB{}, the two sets of parameters yield similar regret on Problem 2, but one is much better than the other on Problem 1. For \MUCB{}, the two sets of parameters yield similar regret on Problem 1, but one is much better than the other on Problem 2. This sheds light on the fact that tuning these parameters may be difficult. On the contrary, the tuning of the Bernoulli GLR test used in \GLRklUCB{} only requires to specify the error probability $\delta$, and setting it to $\delta = 1/\sqrt{T}$ as suggested by Corollary~\ref{cor:Global} yield good performance on both Problem 1 and Problem 2. 


To understand the behavior of the CPD-based algorithms, we analyze their average number of restarts, reported in the tables in Figure~\ref{fig:Pb1} and \ref{fig:Pb2}. In an asymptotic regime (i.e. for $T$ such that Assumption~\ref{ass:LongPeriodsGlobal} or Assumption~\ref{ass:LongPeriods} is satisfied), \GLRklUCB{} should detect all breakpoints with Global restart and all change-points with Local restart. As can be seen, the asymptotic regime is not met in our experiments, except for $T\geq 20000$ on Problem 2 in which \GLRklUCB{} with Global restart performs exactly $\Upsilon_T = 4$ restarts. Besides this case, \GLRklUCB{} typically detects fewer changes than expected, for example between 2 and 3 on Problem 1. Note that \MUCB{} tends to detect fewer changes than \GLRklUCB{}, whereas \CUSUMUCB{} tend to detect more. Especially, when the parameter $\varepsilon$ (giving the minimal amount of change the CUSUM test should detect) is $\varepsilon =0.05$, we observe that \CUSUMUCB{} experiences false-alarms, especially for large horizons (yet this does not prevent the algorithm from having a regret smaller than that of \klUCB{}). Overall, we remark that \GLRklUCB{} is among the best algorithms on both problems for all the horizon values, including the smallest ones: it shows that \GLRklUCB{} is competitive in practice even when the Assumptions~\ref{ass:LongPeriodsGlobal} and~\ref{ass:LongPeriods} are violated.

In these experiments, we tried four variants of \GLRklUCB{}: we investigate the use Global and Local restarts and the use of two exploration sequences: a constant exploration probability $\alpha = \sqrt{{\Upsilon_T A \ln(T)}/{T}}$ and the exploration sequence $\alpha_k = \sqrt{{k A \ln(T)}/{T}}$ that does not require to know the number of breakpoints. We observe that the two types of restarts yield comparable performance (with a slight advantage for Global restarts), and thus investigate the two variants further on a wider benchmark. As for the exploration sequences, we observe that the time-varying one (agnostic to $\Upsilon_T$) always performs best. The reason is that it performs less forced exploration in the first episodes, and as we shall see in our next experiments, scaling down the exploration probability (or exploration sequence) for CPD-based algorithms can lead to better empirical performance. Still, \GLRklUCB{} with a constant exploration probability $\alpha$ also outperforms most of the time other CPD-based algorithms using the exact same $\alpha$.

\paragraph{Robustness on more diverse benchmarks} We now investigate further the performance of the best algorithms for Problem 1 and Problem 2 on a large number of randomly generated piecewise stationary bandit models, with $T=20000$. To generate a random instance, we specify the number of arms $A$, the maximal number of breakpoints $\Upsilon$, a change-point probability $p$, a minimal distance $d_{\min}$, a minimal and maximal amount of change, $\Delta_{\min}$ and $\Delta_{\max}$. Then, we sample the breakpoints uniformly at random under the constraint that $\tau^{k} - \tau^{k-1} \geq d_{\min}$. For each breakpoint $k$, each arm has a probability $p$ to experience a change-point, whose magnitude is chosen uniformly at random in $[\Delta_{\min},\Delta_{\max}]$. 

\begin{figure}[t]

 \centering 
  \includegraphics[height=5.5cm]{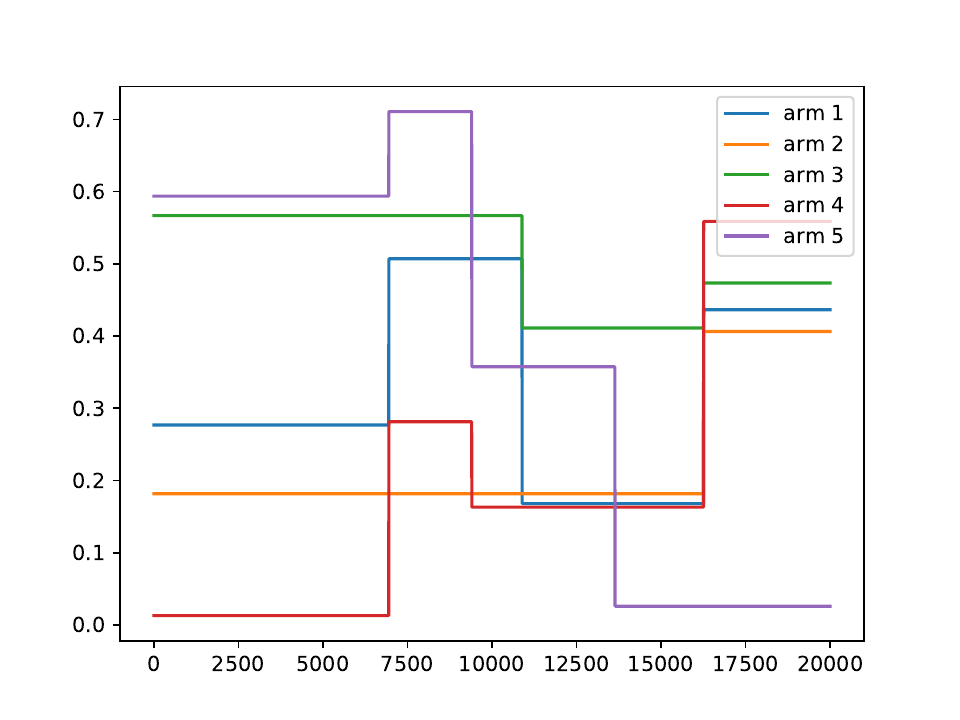}
  \includegraphics[height=5.5cm]{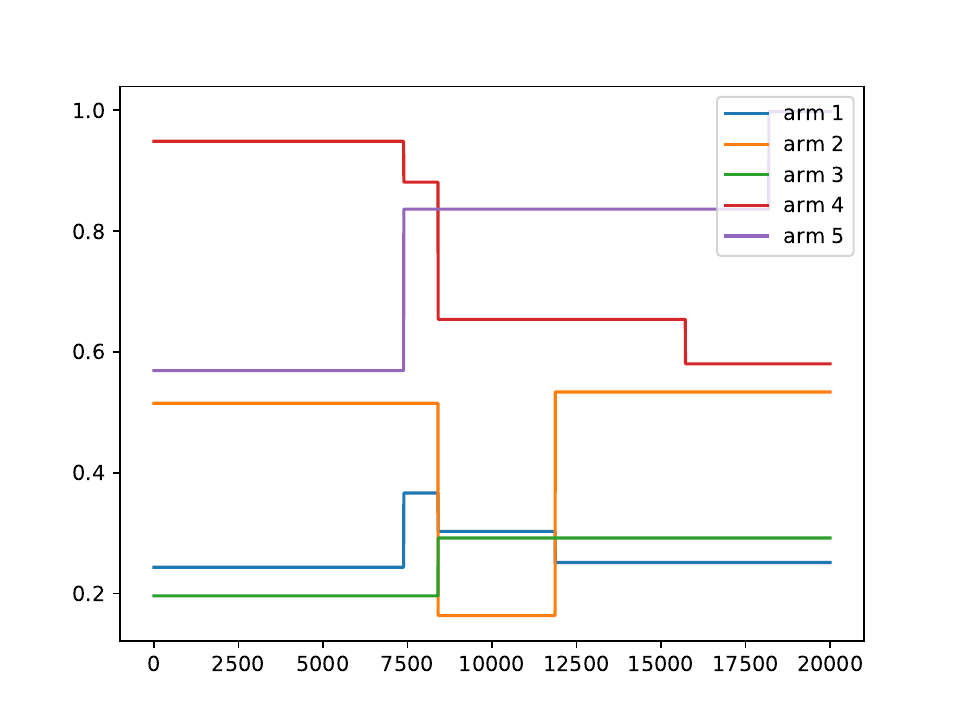}
 
  \includegraphics[height=5cm]{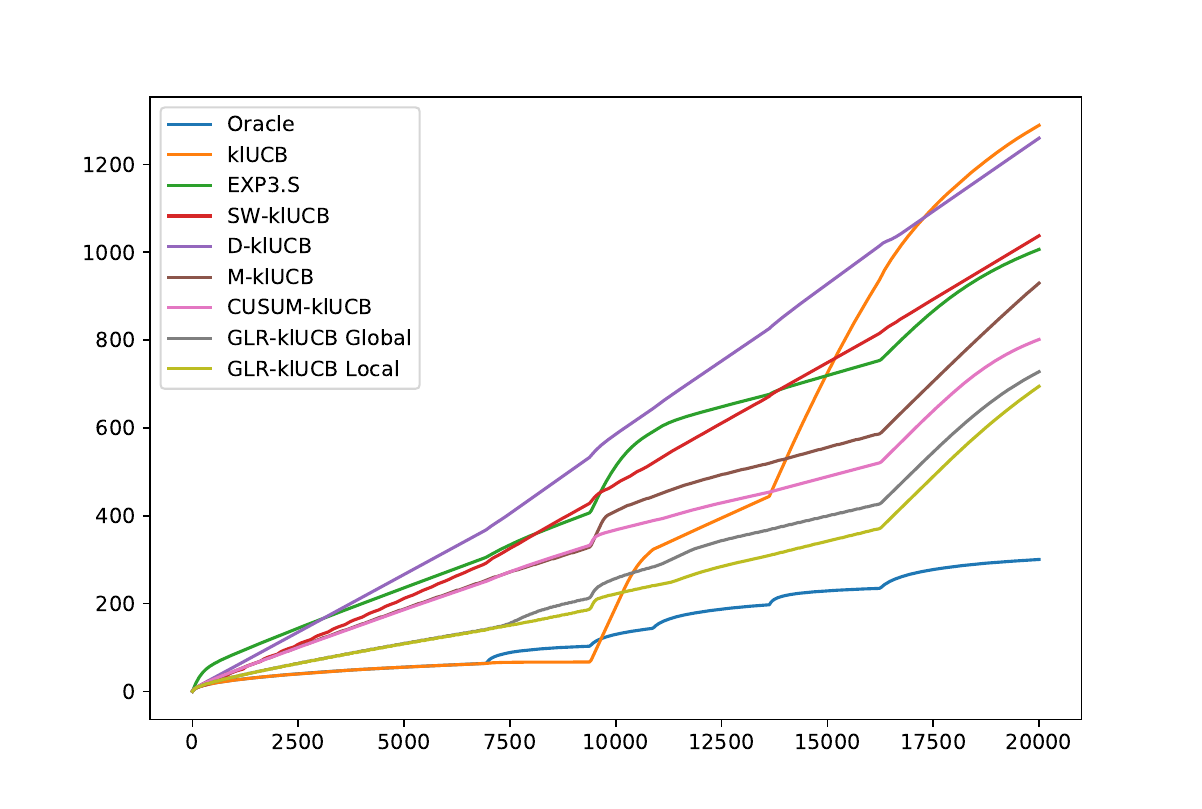}
  \includegraphics[height=5cm]{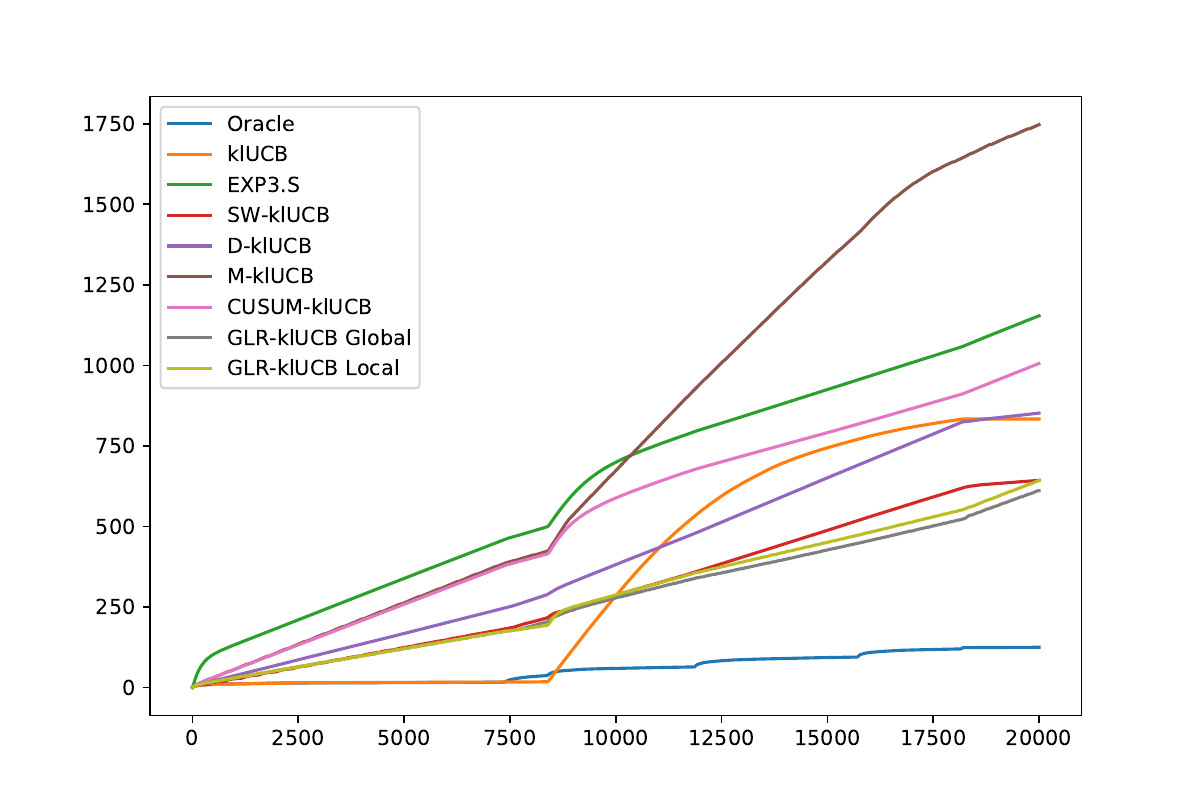}
 {\small
    \begin{center}
    \begin{tabular}{|*{7}{c|}}
    \hline
     Algorithm &  \multicolumn{2}{c|}{Problem 3} & \multicolumn{2}{c|}{Problem 4}  \\
      \hline 
        Oracle-\klUCB{} & $300 \pm 37$ & $5$ & $125 \pm 15$ & $5$ \\
      \hline  
      \klUCB{}  & $1289 \pm 206$  & $0$ & $834 \pm 68$ & $0$ \\
      \hline
      EXP3.S  & $1006 \pm 91$ & $0$  & $1154 \pm 78$ & $0$ \\
      SW-\klUCB{} & $1037 \pm 39$ & $0$ & $\bm{643 \pm 29}$ & $0$ \\
      D-\klUCB{} & $1260 \pm 39$ & $0$ & $852 \pm 31$ & $0$ \\
      \hline
       M-\klUCB{} ($w = 800$)  & $929 \pm 53$ & $1.1$  & $1747 \pm 695$ & $0.5$ \\
      CUSUM-\klUCB{} ($M \!=\! 400, \varepsilon\!=\!0.05$) & $801 \pm 87$ & $20.7$ & $1006 \pm 102$ & $6.277$\\
      \hline
      \GLRklUCB{} Global ($\alpha_k$)& $\bm{728 \pm 65}$ & $3.8$ & $\bm{611 \pm 43}$ & $3.1$ \\
      \GLRklUCB{} Local ($\alpha_k$) & $\bm{695 \pm 59}$ & $4.7$ & $\bm{642 \pm 44}$ & $3.6$ \\
      \hline
    \end{tabular}
    \caption{\label{fig:Pb34}\small 9 algorithms run with $T=20000$ on \textbf{Problem 3} (left) and \textbf{Problem 4} (right): cumulative regret as a function of time (middle row), final regret and number of restarts (bottom row) averaged over $N=1000$ runs.}
\end{center}}

\end{figure}

First, we sample two problems from this procedure with $A=5$ arms, $\Upsilon = 5$ breakpoints, with spacing larger than $d_{\min}=1000$, a magnitude in $[0.05, 0.4]$, and a change-point probability $p=0.5$. Results for these two problems are displayed in Figure~\ref{fig:Pb34}. On \textbf{Problem 3}, there are important changes of the optimal arm as the initial worse arm ends up being the best, with three changes of optimal arms. On \textbf{Problem 4}, the best two arms remain the same but are switched around the middle of the budget $T$. As can be seen in Figure~\ref{fig:Pb34}, on these two instances the two versions of \GLRklUCB{} attain the smallest regret (with SW-\klUCB{} that is very competitive in Problem 4). We note that the Local version performs slightly more restarts (as can already be observed on Problems 1 and 2), but there is again no clear winner between global and local restarts. These experiments also confirm our conclusions regarding the lack of robustness of \MUCB{} (with $w=800$) and \CUSUMUCB{} (with $\varepsilon = \Delta_{\min}$) and their tendency to under-detect or over-detect, respectively.

Finally, in Table~\ref{tab:Bayesian}, we report the regret of the different algorithms averaged over $N=2000$ different instances with $K=5$, $\Upsilon = 6$, $\Delta_{\min}=0.05$, $\Delta_{\max} = 0.3$, $d_{\min} = 1000$ and $p=0.5$. In this last experiment, we also study the influence of using a smaller exploration probability, $\alpha = \alpha_0 \sqrt{{\Upsilon_T A \ln(T)}/{T}}$ with some constant $\alpha_0 \in (0,1]$ (and multiplying $\alpha_k$ by the same $\alpha_0$ for \GLRklUCB{}). We see that for $\alpha_0 = 1$, \GLRklUCB{} outperforms all non-stationary bandit algorithms, but not \klUCB{}. This can be explained by the fact that some of the random instances may contain no change of optimal arm, and \klUCB{} is expected to be good in this setting, without paying the price of forced exploration. However, when we reduce the amount of forced exploration by setting $\alpha_0$ to 0.5 and 0.1, \klUCB{} is outperformed by actively adaptive algorithms, out of which \GLRklUCB{} has the best performance, with a slight advantage for Global restart.  

\begin{table}[h]
{\small
    \begin{center}
    \begin{tabular}{|*{7}{c|}}
    \hline
     \textbf{Algorithm} &  \multicolumn{2}{c|}{$\alpha_0 = 1$} & \multicolumn{2}{c|}{$\alpha_0 = 0.5$} & \multicolumn{2}{c|}{$\alpha_0 = 0.1$}   \\
      \hline 
      \klUCB{}  & $\bm{515}$ & $0$ & $515$ & $0$ & $515$ & $0$\\
      \hline
      EXP3.S  & $1083$ & $0$ & $1083$ & $0$ &$1083$ & $0$  \\
      SW-\klUCB{} & $826$ & $0$ & $826$ & $0$ & $826$ & $0$ \\
      D-\klUCB{} & $1022$ & $0$ & $1022$ & $0$ & $1022$ & $0$  \\
      \hline
      M-\klUCB{} ($w = 800$)  & $1070$ & $0.9$ & $716$ & $0.8$ & $415$ & $0.8$ \\
      CUSUM-\klUCB{} ($M \!=\! 400, \varepsilon\!=\!0.05$) & $973$ & $12.4$ & $640$ & $10.9$ & $347$ & $10.9$ \\
      \hline
      \GLRklUCB{} Global ($\alpha_k$)& $\bm{611}$ & $3.1$ & $ \bm{404}$ & $3.0$ & $\bm{252}$ & $2.9$ \\
      \GLRklUCB{} Local ($\alpha_k$) & $\bm{691}$ & ${4.4}$ & $\bm{462}$ & ${3.8}$ & $\bm{284}$ & $3.5$ \\
      \hline
    \end{tabular}
    \caption{\small \label{tab:Bayesian}Regret and number of restarts for several algorithm run with $T=20000$, averaged over $N=20000$ random problem instances with $K=5$, $\Upsilon = 6$, $\Delta_{\min}=0.05$, $\Delta_{\max} = 0.3$, $d_{\min} = 1000$ and $p=0.5$. 
    }
\end{center}}
 
\end{table}

\paragraph{GLR-klUCB beyond piecewise stationary models}
We summarize in this paragraph the experimental results of \citep{Seznec20RestlessRotting}, who performed experiments including \GLRklUCB{} on a restless rotting bandit problem, in which the reward function $\mu_a(t)$ of each arm is assumed to be a non-increasing function of $t$. They designed 9 bandit games with arms mean rewards learned from the R6A - Yahoo! Front Page Today Module. In this setup, each arm corresponds to a news article and its mean reward at a certain time step (the probability that the news is clicked on) is learned from the dataset with a sliding window average of 30 000 samples. Each game corresponds to a 12 hours timeframe between 6 p.m. and 6 a.m. EST during one day of May 2009 and the number of rounds is equal to the number of users visiting the Yahoo! Front Page on the period. This time frame was chosen because the mean rewards (click probabilities) are mostly decaying. However, besides this high-level data selection, they do not enforce any of their theoretical assumptions (the reward functions are neither piecewise stationary, nor strictly non-increasing).

They studied \GLRklUCB{} (with Gaussian confidence intervals) with no forced exploration ($\alpha =0$) and local restart together with Exp3.S and two of their algorithms designed for the rotting case, namely RAW-UCB and FEWA. \GLRklUCB{} recovers consistently the best performance on the 9 games, significantly outperforming FEWA and Exp3.S. RAW-UCB performs almost the same as \GLRklUCB{}, though the former does not need the knowledge of $T$ to be tuned. Interestingly, the regret of \GLRklUCB{} has the same logarithmic shape as FEWA or RAWUCB when one arm is significantly above the others. This is not the case for Exp3.S which keeps a small linear trend in regret due to random exploration.  These algorithms were proved to have logarithmic regret on each stationary part of a non-increasing piecewise bandit problem. This suggests that \GLRklUCB{} with no forced exploration and local restart could also enjoy logarithmic regret in a similar setting.


\section{Conclusion}

We proposed a new algorithm based on a change-point detector which empirically outperforms other CPD-based approaches designed for the general piecewise-stationary bandit problems, while attaining similar regret without the knowledge of the number of breakpoints and without any prior knowledge of the arms means. We proved that \GLRklUCB{} attains a $\mathcal{O}(\sqrt{\Upsilon_T A T \log(T)}/(\Delta^{\text{change}})^2)$ regret for ``easy'' instances in which the breakpoints are far enough from each other. When the smallest magnitude of a breakpoint $\Delta^{\text{change}}$ is not too small, this is comparable to the regret of recently proposed algorithms that are agnostic to $\Upsilon_T$ but whose implementation is much more intricate than that of \GLRklUCB{}. 

The presence of $\Delta^{\text{change}}$ in the regret bound comes from the fact that our analysis (as that of other CPD-based algorithms) assumes that all the breakpoints are detected by the algorithm. But our experiments reveal that on ``harder'' instances, the algorithm actually does not detect all the breakpoints and still attain small regret. Hence, in future work, we intend to work on an improved analysis of \GLRklUCB{} introducing some notion of ``meaningful changes'' that need to be detected by the algorithm to ensure a small regret. We will also investigate whether \GLRklUCB{} can be used without forced exploration under some extra assumption on the types of breakpoints encountered, such as global changes \citep{OdalricSubo19} or restless rotting bandits \citep{Seznec20RestlessRotting}.

\acks{The research presented was supported by European CHIST-ERA project DELTA and the French National Research Agency project BADASS (ANR-16-CE40-0002) and BOLD (ANR-19-CE23-0026-04).}
%

\bibliography{biblio}


\appendix
\onecolumn

\section{Proof of Proposition~\ref{prop:EnoughSamples}}\label{proof:EnoughSamples}

We consider one arm $a\in\{1,\dots,A\}$, and when the \GLRklUCB{} algorithm is running,
we consider two time steps $s\leq t\in\N^\star$, chosen between two restart times for that arm $a$. All time steps $u \in \{s,\dots,t\}$  belong to the same episode, that is $k_u=k_t$.
Lines~$3$-$4$ of Algorithm~\ref{algo:GLRklUCB} imply that for all $u \in \{s+1,\dots,t\}$, 
\[\left\{u\mod \left\lceil \frac{A}{\alpha_{k_t}} \right\rceil = a\right\}\subset \{A_u = a\}\]
Thus we have
\begin{align*}
    n_a(t) - n_a(s)
    & = \sum_{u=s+1}^{t} \ind(A_u = a) \\
      & \geq \sum_{u=s+1}^{t} \ind\left(u\mod \left\lceil \frac{A}{\alpha_{k_t}} \right\rceil = a\right) \\
    & = \bigl(t - (s+1) + 1\bigr) / \left\lceil \frac{A}{\alpha_{k_t}} \right\rceil
    \geq \left\lfloor \frac{\alpha_{k_t}}{A} (t - s) \right\rfloor,
\end{align*}
which proves Proposition~\ref{prop:EnoughSamples}.


\section{Concentration Inequalities}\label{proof:Conc}



\subsection{Proof of Lemma~\ref{lem:FalseAlarm}} \label{proof:FalseAlarm}

Lemma~\ref{lem:FalseAlarm} is presented for bounded distributions and is actually valid for any sub-Bernoulli distribution. It could also be presented for more general distributions satisfying
\begin{equation}
 \bE[e^{\lambda X} ] \leq e^{\phi_\mu(\lambda)} \ \ \text{ with } \ \ \mu=\bE[X]\label{def:subExpo},
\end{equation}
where $\phi_\mu(\lambda)$ is the log moment generating of some one-dimensional exponential family. The Bernoulli divergence $\kl(x,y)$ would be replaced by the corresponding divergence in that exponential family (which is the Kullback-Leibler divergence between two distributions of means $x$ and $y$).

    Let's go back to the Bernoulli case with divergence given in \eqref{BernoulliDivergence}. A first key observation is
    \[s \times \kl\left(\hat{\mu}_{1:s},\hat{\mu}_{1:n}\right) + (n-s) \times \kl\left(\hat{\mu}_{s+1:n},\hat{\mu}_{1:n}\right) = \inf_{\mu \in [0,1]}\left[s \times \kl\left(\hat{\mu}_{1:s},\lambda\right) + (n-s) \times \kl\left(\hat{\mu}_{s+1:n},\lambda\right)\right].\]
    Hence the probability of a false alarm occurring is upper bounded as
    \begin{align*}
        \bP_{\mu_0}\left(T_{\delta} < \infty\right) & \leq \bP_{\mu_0}\left(\exists (s,n) \in \N^2, s < n:  s \, \kl\left(\hat{\mu}_{1:s},\hat{\mu}_{1:n}\right) + (n-s) \, \kl\left(\hat{\mu}_{s+1:n},\hat{\mu}_{1:n}\right) > \beta(n,\delta)\right)\\
        & \leq \bP_{\mu_0}\left(\exists (s,n) \in \N^2, s < n:  s \, \kl\left(\hat{\mu}_{1:s},\mu_0\right) + (n-s) \, \kl\left(\hat{\mu}_{s+1:n},\mu_0\right) > \beta(n,\delta)\right) \\
        & \leq \sum_{s =1}^{\infty} \bP_{\mu_0}\left(\exists n > s:  s \, \kl\left(\hat{\mu}_{1:s},\mu_0\right) + (n-s) \, \kl\left(\hat{\mu}_{s+1:n},\mu_0\right) > \beta(n,\delta)\right)\\
        & =   \sum_{s =1}^{\infty} \bP_{\mu_0}\left(\exists r \in \N:  s \, \kl\left(\hat{\mu}_{s},\mu_0\right) + r \, \kl\left(\hat{\mu}'_{r},\mu_0\right) > \beta(s+r,\delta)\right),
    \end{align*}
    where $\hat{\mu}_{s}$ and $\hat{\mu}'_{r}$ are the empirical means of respectively $s$ and $r$ \iid{} observations with mean $\mu_0$ and distribution $\nu$, that are independent from the previous ones.
    As $\nu$ is sub-Bernoulli, the conclusion follows from Lemma~\ref{lem:ConcFirst} below and from the definition of $\beta(n,\delta)$:

    \begin{align*}
        &\bP_{\mu_0}\left(T_{\delta} < \infty\right) \\
        & \leq \sum_{s =1}^{\infty} \bP_{\mu_0}\left(\exists r \in \N^\star : s \, \kl\left(\hat{\mu}_{s},\mu_0\right) + r \, \kl\left(\hat{\mu}'_{r},\mu_0\right) > 6\ln(1+\ln(s+r))+ 2 \cT\left(\frac{\ln(3(s+r)^{3/2}/\delta)}{2}\right)\right)\\
        & \leq \sum_{s =1}^{\infty} \bP_{\mu_0}\left(\exists r \in \N^\star : s \, \kl\left(\hat{\mu}_{s},\mu_0\right) + r \, \kl\left(\hat{\mu}'_{r},\mu_0\right) > 3\ln(1+\ln(s))+3\ln(1+\ln(r)) + 2 \cT\left(\frac{\ln(3s^{3/2}/\delta)}{2}\right)\right)
    \end{align*}
    And so we have $\bP_{\mu_0}\left(T_{\delta} < \infty\right) \leq \sum_{s =1}^{\infty} \frac{\delta}{3s^{3/2}} \leq \delta$.

    \begin{lemma}\label{lem:ConcFirst}
    $(X_i)_{i \in \N}$ and $(Y_i)_{i \in \N}$ two independent \iid{} processes with resp. means $\mu$ and $\mu'$ such that
    \[\bE[e^{\lambda X_1}] \leq e^{\phi_{\mu}(\lambda)} \ \ \text{and} \ \ \bE[e^{\lambda Y_1}] \leq e^{\phi_{\mu'}(\lambda)},\]
    where $\phi_\mu(\lambda) = \bE_{X \sim \nu^{\mu}}[e^{\lambda X}]$ is the moment generating function of the distribution $\nu^\mu$, which is the unique distribution in an exponential family that has mean $\mu$.
    Let $\kl(\mu,\mu') = \KL(\nu^\mu,\nu^{\mu'})$ be the divergence function associated to that exponential family. Introducing the notation $\hat{\mu}_s = \frac{1}{s}\sum_{a=1}^s X_i$ and $\hat{\mu}'_r = \frac{1}{r}\sum_{a=1}^r Y_i$, it holds that for every $s,r \in \N^\star$,
    \[\bP\left(\exists r \in \N^\star : s \, \kl\left(\hat{\mu}_{s},\mu\right) + r \, \kl\left(\hat{\mu}_r',\mu'\right) > 3\ln(1+\ln(s)) + 3\ln(1+\ln(r)) + 2\cT\left(\frac{x}{2}\right)\right)\leq e^{-x},\]
    where $\cT$ is the function defined in \eqref{def:function_T}.
\end{lemma}

\paragraph{Proof of Lemma~\ref{lem:ConcFirst}}
Using the same construction as in the proof of Theorem~14 in \cite{KK18Martingales}, one can prove that for every $\lambda \in I$ (for an interval $I$), there exists a non-negative super-martingale $M^\lambda (s)$ with respect to the filtration $\cF_t = \sigma(X_1,\dots,X_t)$ that satisfies $\bE[M^\lambda(s)] \leq 1$ and
\[\forall s \in \N^\star, \ \ M^\lambda (s) \geq e^{\lambda [s\kl(\hat{\mu}_s,\mu) - 3\ln(1+\ln(s))] - g(\lambda)}\]
for some function $g : I \rightarrow \R$. This super-martingale is of the form
\[M^\lambda (s) = \int e^{\eta\sum_{a=1}^s X_i - \phi_\mu(\lambda)s} d\pi(\eta)\]
for a well-chosen probability distribution $\pi$, and the function $g$ can be chosen to be any
\begin{small}
\begin{eqnarray*}
    g_{\xi} : \left[0; 1/(1+\xi)\right] & \longrightarrow & \R \\
    \lambda & \mapsto & \lambda(1+\xi) \ln \left(\frac{\pi^2}{3(\ln(1+\xi))^2}\right) -  \ln(1 - \lambda(1 + \xi))
\end{eqnarray*}
\end{small}
for a parameter $\xi \in [0,1/2]$.

Similarly, there exists an independent super-martingale $W^\lambda (r)$ w.r.t. the filtration $\cF_r' = \sigma(Y_1,\dots,Y_r)$ such that
\[\forall r \in \N^\star, \ \ W^\lambda (r) \geq e^{\lambda [r\kl(\hat{\mu}'_r,\mu) - 3\ln(1+\ln(r))] - g(\lambda)},\]
for the same function $g(\lambda)$. In the terminology of \cite{KK18Martingales}, the processes $\bm{X}(s) = s \, \kl(\hat{\mu}_s,\mu) - 3\ln(1+\ln(s))$ and $\bm{Y}(s) = r \, \kl(\hat{\mu}_r,\mu) - 3\ln(1+\ln(r))$ are called $g$-DCC for Doob-Cram\'er-Chernoff, as Doob's inequality can be applied in combination with the Cram\'er-Chernoff method to obtain deviation inequalities that are uniform in time.

Here we have to modify the technique used in their Lemma 4 in order to take into account the two stochastic processes, and the presence of super-martingales instead of martingales (for which Doob inequality still works). One can write
\begin{align*}
    &\bP\left(\exists r \in \N^\star : s \, \kl\left(\hat{\mu}_{s},\mu\right) + r \, \kl\left(\hat{\mu}_r',\mu'\right) > 3\ln(1+\ln(s)) + 3\ln(1+\ln(r)) + u\right) \\
    & \leq \bP\left(\exists r \in \N^\star : M^\lambda(s)W^\lambda(r) > e^{\lambda u - 2g(\lambda)}\right) \\
    & = \lim_{n \rightarrow \infty} \bP\left(\exists r \in \{1,\dots,n\} :  M^\lambda(s)W^\lambda(r) > e^{\lambda u - 2g(\lambda)}\right) \\
    & = \lim_{n \rightarrow \infty} \bP\left(\sup_{r \in \{1,\dots,n\}}  M^\lambda(s)W^\lambda(r) > e^{\lambda u - 2g(\lambda)}\right).
\end{align*}
Using that $\tilde{M}(r) = M^\lambda(s)W^\lambda(r)$ is a super-martingale with respect to the filtration \[\tilde{\cF}_r = \sigma(X_1,\dots,X_s, Y_1,\dots,Y_r),\] one can apply Doob's maximal inequality to obtain
\begin{eqnarray*}
    \bP\left(\sup_{r \in \{1,\dots,n\}}  M^\lambda(s)W^\lambda(r) > e^{\lambda u - 2g(\lambda)}\right) &\leq& e^{-(\lambda u - 2g(\lambda))}\bE[\tilde{M}(1))] \\
    & = & e^{-(\lambda u - 2g(\lambda))}\bE[M^\lambda(s)W^\lambda(1)] \\
    & \leq & e^{-(\lambda u - 2g(\lambda))},
\end{eqnarray*}
using that $M^\lambda(s)$ and $W^\lambda(1)$ are independent and have an expectation smaller than $1$.

Putting things together yields
\[\bP\left(\exists r \in \N^\star : s \, \kl\left(\hat{\mu}_{s},\mu\right) + r \, \kl\left(\hat{\mu}_r',\mu'\right) > 3\ln(1+\ln(s)) + 3\ln(1+\ln(r)) + u\right) \leq e^{-\left(\lambda u - 2g_{\xi}(\lambda)\right)},\]
for any function $g_{\xi}$ defined above.
The conclusion follows by optimizing for both $\lambda$ and $\xi$, using Lemma~18 in \cite{KK18Martingales}.

\subsection{A Concentration Result Involving Two Arms}\label{proof:Chernoff2arms}

The following result is useful to control the probability of the good event in our two regret analyzes. Its proof follows from a straightforward application of the Cram\'er-Chernoff method \citep{Boucheron2013}.

\begin{lemma}\label{lem:Chernoff2arms}
    Let $\hat{\mu}_{i,s}$ be the empirical mean of $s$ \iid{} observations with mean $\mu_i$, for $i \in \{a,b\}$, that are $\sigma^2$-sub-Gaussian. Define $\Delta = \mu_a - \mu_b$. Then for any $s,r > 0$, we have
    \begin{eqnarray*}
        \bP\left(\frac{sr}{s+r}\Big(\hat{\mu}_{a,s} - \hat{\mu}_{b,r} - \Delta\Big)^2 \geq u \right) &\leq& 2\exp\left(- \frac{u}{2\sigma^2}\right)\cdot
    \end{eqnarray*}
\end{lemma}


\paragraph{Proof of Lemma~\ref{lem:Chernoff2arms}}

We first note that
\begin{align}\label{Preparation}
	&\bP\left(\frac{sr}{s+r}\Big(\hat{\mu}_{a,s} - \hat{\mu}_{b,r} - \Delta\Big)^2 \geq u \right) \nonumber\\
	&\leq \bP\left(\hat{\mu}_{a,s} - \hat{\mu}_{b,r} \geq \Delta + \sqrt{\frac{s+r}{sr}u} \right) + \bP\left(\hat{\mu}_{b,r} - \hat{\mu}_{a,s} \geq -\Delta + \sqrt{\frac{s+r}{sr}u} \right),
\end{align}
and those two quantities can be upper-bounded similarly using the Cram\'er-Chernoff method.

Let $(X_i)$ and $(Y_i)$ be two \iid{} sequences that are $\sigma^2$ sub-Gaussian with mean $\mu_1$ and $\mu_2$ respectively. Let $n_1$ and $n_2$ be two integers and $\hat\mu_{1,n_1}$ and $\hat{\mu}_{2,n_2}$ denote the two empirical means based on $n_1$ observations from $X_i$, and $n_2$ observations from $Y_i$ respectively.
Then for every $\lambda > 0$, we have
\begin{eqnarray*}
 \bP\left(\hat\mu_{1,n_1} - \hat\mu_{2,n_2} \geq \mu_1 - \mu_2 + x \right) & \leq & \bP\left(\frac{1}{n_1}\sum_{a=1}^{n_1} (X_i-\mu_1) - \frac{1}{n_2}\sum_{a=1}^{n_2} (Y_i - \mu_2) \geq x \right)\\
 & \leq & \bP\left(e^{\lambda \left(\frac{1}{n_1}\sum\limits_{i=1}^{n_1} (X_i-\mu_1) - \frac{1}{n_2}\sum\limits_{i=1}^{n_2} (Y_i - \mu_2)\right)} \geq e^{\lambda x} \right) \\
 \text{(using Markov's inequality)}
 & \leq & e^{-\lambda x}\bE\left[e^{\lambda \frac{1}{n_1}\sum\limits_{i=1}^{n_1} (X_i-\mu_1)}\right]\bE\left[e^{-\lambda \frac{1}{n_2}\sum\limits_{i=1}^{n_2} (Y_i-\mu_2)}\right] \\
 & = & \exp\left(- \lambda x + n_1\phi_{X_1 - \mu_1}\left(\frac{\lambda}{n_1}\right) + n_2\phi_{Y_1 - \mu_2}\left(-\frac{\lambda}{n_2}\right)\right) \\
 & \leq & \exp\left(- \lambda x + \frac{\lambda^2\sigma^2}{2n_2} + \frac{\lambda^2\sigma^2}{2n_1}\right),
\end{eqnarray*}
where the last inequality uses the sub-Gaussian property.
Choosing the value $\lambda = \frac{x}{2\left[\sigma^2/(2n_1) + \sigma^2/(2n_2)\right]}$ which minimizes the right-hand side of the inequality yields
\[\bP\left(\hat\mu_{1,n_1} - \hat\mu_{2,n_2} \geq \mu_1 - \mu_2 + x \right) \leq \exp\left(- \frac{n_1n_2}{n_1+n_2}\frac{x^2}{2\sigma^2}\right).\]
Using this inequality twice in the right hand side of \eqref{Preparation} concludes the proof.

\section{Elements of the analysis of \GLRklUCB{} with Global Restarts}\label{app:Global}

We present in this section the detailed proof of the two crucial lemmas in the analysis of \GLRklUCB{} with Global Restarts.

\subsection{Proof of Lemma~\ref{lem:BanditGlobal}}

Lemma~\ref{lem:BanditGlobal} follows from summing over $k$ and $a$ upper bounds on the quantities 
\[E_{k,a} : = \bE\left[\ind(\cE_T) \sum_{t={\tau}^{(k)}+1}^{\tau^{(k+1)}} \Delta_{a}^{(k)} \ind{\left(A_t = a, \UCB_{a}(t-1) \geq \UCB_{k^\star}(t-1)\right)} \right],\]
for each $k \in \{0,\dots,\Upsilon_T\}$ and each arm $a$ such that $\Delta_a^{(k)} > 0$. 

Using that on $\cE_T$, $\hat\tau^{(k)} \leq \tau^{(k)} + d^{(k)}$, one can write $E_{k,a} \leq   \Delta_{a}^{(k)}d^{(k)} + R_{k,a}$ where
{\small
\begin{align*}
 R_{k,a} & =\bE\left[\ind(\cE_T)\sum_{t={\hat\tau}^{(k)}+1}^{\tau^{(k+1)}}\Delta_{a}^{(k)} \ind{\left(A_t = a,\UCB_{a}(t-1) \geq \UCB_{k^\star}(t-1)\right)} \right] \\
 & = \underbrace{\bE\left[\ind(\cE_T)\!\!\!\sum_{t={\hat\tau}^{(k)}+1}^{\tau^{(k+1)}}\!\!\! \ind{\left(\UCB_{k^\star}(t-1) \leq \mu_{k^\star}^{(k)}\right)} \right]}_{:=(A)}  +  \underbrace{\bE\left[\ind(\cE_T)\!\!\!\sum_{t={\hat\tau}^{(k)}+1}^{\tau^{(k+1)}} \Delta_{a}^{(k)}\ind{\left(A_t = a, \UCB_{a}(t-1) \geq \mu_{k^\star}^{(k)}\right)} \right]}_{:=(B)}
\end{align*}}

To upper bound $(A)$ and $(B)$, we recall the following notation. We let $\hat\tau(t)$ be the last time before $t$ that the algorithm restarted. Moreover, we denote by $n_a(t) = \sum_{s=\tau(t)+1}^{t}\ind(A_s = a)$ the number of selections of arm $a$ since the last (global) restart, and $\hat{\mu}_a(t) = \frac{1}{n_a(t)}\sum_{s=\tau(t)+1}^{t}X_{a,s}\ind(A_s = a)$ their empirical average (if $n_a(t) \neq 0$).

\paragraph{Upper bound on Term $\bm{(A)}$} By definition of the Upper Confidence Bound, one can write

\begin{align*}
    (A) & \leq \bE\left[\ind(\cE_T) \sum_{t=\hat \tau^{(k)}}^{\tau^{(k+1)-1}} \ind\left(n_{k^\star}(t) \, \kl\left(\hat{\mu}_{k^\star}(t), \mu_{k^\star}^{(k)}\right) \geq f(t - \hat\tau(t))\right)\right] \\
    & \leq \bE\left[\ind(\cC^{(k)}) \sum_{t=\hat{\tau}^{(k)}}^{\tau^{(k+1)}-1}\ind\left(n_{k^\star}(t) \, \kl\left(\hat{\mu}_{k^\star}(t), \mu_{k^\star}\right) \geq f(t - \hat\tau^{(k)})\right)\right],
\end{align*}
where we introduce the event $\cC^{(k)}$ that all the changes up to the $k$-th have been detected:
\begin{equation}\label{def:EventCiGlobal}
    \cC^{(k)} = \left\{\forall j \leq k, \hat\tau^{(j)} \in \{\tau^{(j)} + 1, \dots, \tau^{(j)} + d^{(j)} \} \right\}.
\end{equation}
Clearly, $\cE_T \subseteq \cC^{(k)}$ and $\cC^{(k)}$ is $\cF_{\hat\tau^{(k)}}$-measurable. Observe that conditionally to $\cF_{\hat\tau^{(k)}}$, when $\cC^{(k)}$ holds, $\hat{\mu}_{k^\star}(t)$ is the average of samples that have all mean $\mu_{k^\star}^{(k)}$.
Thus, introducing $\hat{\mu}_s$ as a sequence of \iid{} random variables with mean $\mu_{k^\star}^{(k)}$, one can write
\begin{align*}
    &\bE\left[\left.\ind(\cC^{(k)}) \sum_{t=\hat{\tau}^{(k)}}^{\tau^{(k+1)}} \ind\left(n_{k^\star}(t) \, \kl\left(\hat{\mu}_{k^\star}(t), \mu_{k^\star}^{(k)}\right) \geq f\left(t - \hat\tau^{(k)}\right)\right) \right| \cF_{\hat\tau^{(k)}}\right]\\
    & = \ind(\cC^{(k)}) \sum_{t=\hat{\tau}^{(k)}}^{\tau^{(k+1)}} \bE\left[\left.\ind\left(n_{k^\star}(t) \, \kl\left(\hat{\mu}_{k^\star}(t), \mu_{k^\star}^{(k)}\right) \geq f\left(t - \hat\tau^{(k)}\right)\right) \;\right|\; \cF_{\hat\tau^{(k)}}\right] \\
    & \leq \ind(\cC^{(k)}) \sum_{t'=1}^{\tau^{(k+1)} - \hat{\tau}^{(k)}} \bP\left(\exists s \leq t' : s \, \kl\left(\hat{\mu}_{s},\mu_{k^\star}^{(k)}\right) \geq f(t')\right). 
\end{align*}
Using the concentration inequality given in Lemma 2 of \cite{KLUCBJournal} and the fact that $f(t) = \ln(t) + 3 \ln(\ln(t))$ allow to upper bound the probability corresponding to term $t'$ by $1/(t'\ln(t'))$. Using the law of total expectation yields 
\[(A)  \leq 2+\sum_{t=3}^{\tau^{(k+1)} - {\tau}^{(k)}} \frac{1}{t\ln(t)} \leq 3+\ln\left(\ln\left(\tau^{(k+1)} - \tau^{(k)}\right)\right).\]

\paragraph{Upper bound on Term $\bm{(B)}$} We let $\tilde{\mu}_{a,s}^{(k)}$ denote the empirical mean of the first $s$ observations of arm $a$ made after time $t=\hat{\tau}^{(k)}+1$. Rewriting the sum in $t$ as the sum of consecutive intervals $[\tau^{(k)}+1, \tau^{(k+1)}]$,
\begin{align*}
    (B) & = \Delta_a^{(k)} \bE\Big[\ind(\cE_T) \sum_{t=\hat\tau^{(k)}+1}^{\tau^{(k+1)}}\ind\left(A_t=a,\UCB_{a}(t-1) \geq \mu_{k^\star}^{(k)}\right)\Big] \\
    &= \Delta_a^{(k)} \bE\Big[\ind(\cE_T) \sum_{t=\hat\tau^{(k)}+1}^{\tau^{(k+1)}}\ind\left(A_t=a,n_a(t-1)\kl\left(\hat\mu_{a}(t-1) , \mu_{k^\star}^{(k)}\right) \leq f\left(t - \hat\tau^{(k)}\right)\right)\Big] \\
     & \leq  \Delta_a^{(k)} \bE\left[\ind(\cE_T) \sum_{t=\hat{\tau}^{(k)}+1}^{\tau^{(k+1)}}\sum_{s=1}^{t-\hat\tau^{(k)}}\ind\left(A_t = a,n_a(t-1) = s\right) \ind\left(s \, \kl(\tilde{\mu}_{a,s}^{(k)}, \mu_{k^\star}^{(k)}) \leq f\left(\tau^{(k+1)} - \tau^{(k)}\right)\right)\right] \\
    & \leq \Delta_a^{(k)}\bE\Big[\ind(\cE_T) \sum_{s=1}^{n_a(\tau^{(k+1)})}\ind\left(s \, \kl(\tilde{\mu}_{a,s}^{(k)}, \mu_{k^\star}^{(k)}) \leq f(\tau^{(k+1)} - \tau^{(k)})\right)\Big]  \\
        & \leq \Delta_a^{(k)}\bE\Big[\ind(\cC_k) \sum_{s=1}^{n_a(\tau^{(k+1)})}\ind\left(s \, \kl(\tilde{\mu}_{a,s}^{(k)}, \mu_{k^\star}^{(k)}) \leq f(\tau^{(k+1)} - \tau^{(k)})\right)\Big],  
\end{align*}
where $\cC^{(k)}$ is the event already defined in \eqref{def:EventCiGlobal}. Conditionally to $\cF_{\hat\tau^{(k)}}$, when $\cC^{(k)}$ holds, for $s \in \{1, \ldots, n_a(\tau^{(k+1)})\}$, $\tilde{\mu}_{a,s}^{(k)}$ is the empirical mean from \iid{} observations of mean $\mu_a^{(k)}$.
Therefore, introducing $\hat{\mu}_s$ as a sequence of \iid{} random variables with mean $\mu_a^{(k)}$, it follows from the law of total expectation that
\begin{align*}
    (B)& \leq\Delta_a^{(k)} \sum_{s=1}^{\tau^{(k+1)}-\tau^{(k)}}\bP\left(s \times \kl(\hat{\mu}_{s}, \mu_{k^\star}^{(k)}) \leq f\left(\tau^{(k+1)} - \tau^{(k)}\right)\right)\cdot
\end{align*}
As $\hat{\mu}_s$ is the empirical mean of i.i.d. observation of mean $\mu_a^{(k)}$  and $\mu_{k^\star}^{(k)} > \mu_a^{(k)}$, and upper bound on this sum of probabilities can be found in Appendix A.2  of \cite{KLUCBJournal}, which yields
\[
    (B)  \leq  \frac{\Delta_a^{(k)}}{\kl(\mu_a^{(k)},\mu_{k^\star}^{(k)})}\ln\left(\tau^{(k+1)} - \tau^{(k)}\right) + \bigO{\sqrt{\ln\left(\tau^{(k+1)} - \tau^{(k)}\right)}}\;.\]

\paragraph{Conclusion} Combining the upper bound on (A) and (B) yields
\[E_{k,a} \leq \Delta_{a}^{(k)}d^{(k)} + \frac{\Delta_a^{(k)}\ln\left(\tau^{(k+1)}-\tau^{(k)}\right)}{\kl\left(\mu_a^{(k)},\mu_{k^\star}^{(k)}\right)} + \bigO{\sqrt{\ln\left(\tau^{(k+1)}-\tau^{(k)}\right)}}\;.\]
The bound in Lemma~\ref{lem:BanditGlobal} follows from the observation that $E_{k,a}$ is also trivially upper bounded by $\Delta_a^{(k)}\left(\tau^{(k+1)}-\tau^{(k)}\right)$.

\subsection{Proof of Lemma~\ref{lem:GoodEventGlobal}}\label{app:CrucialLemma}


Recall that $\cC^{(k)}$ defined in \eqref{def:EventCiGlobal} is the event that all the breakpoints up to the $k$-th have been correctly detected.
Using a union bound, one can write
\begin{align*}
    \bP(\cE_T^c) & \leq \sum\limits_{k=1}^{\Upsilon_T}
    \bP\left(\left. \hat\tau^{(k)} \notin \{ \tau^{(k)}+1, \dots, \tau^{(k)} + d^{(k)} \} \right|\; \cC^{(k-1)}\right)  + \bP\left(\left.\hat\tau^{(\Upsilon_T+1)} \leq T \right| \cC^{(\Upsilon_T)}\right) \\
    \\ & \leq \sum\limits_{k=1}^{\Upsilon_T+1} \underbrace{\bP\left(\hat\tau^{(k)} \leq \tau^{(k)} \;|\; \cC^{(k-1)}\right)}_{(a)} + \sum\limits_{k=1}^{\Upsilon_T} \underbrace{\bP\left(\hat\tau^{(k)} \geq \tau^{(k)} + d^{(k)} \;|\; \cC^{(k-1)}\right)}_{(b)}.
\end{align*}
The final result follows by proving that $(a) \leq A\delta$ and $(b)\leq \delta$, as detailed below.

\paragraph{Upper bound on $(a)$: controlling the false alarm}
$\hat\tau^{(k)} \leq \tau^{(k)}$ implies that there exists an arm whose associated change point detector has experienced a false-alarm. Under the bandit algorithm, the change point detector associated with each arm $a$ is based on (possibly much) less than $t - \tau_a(t)$ samples from arm $a$, which makes false alarm even less likely to occur. More precisely, we upper bound term $(a)$ by 
\begin{align*}
    (a) & \leq \bP\left(\exists a, \exists s < t \leq n_a(\tau^{(k)}) : s \, \kl\left(\tilde{\mu}_{a,1:s}^{(k-1)},\tilde{\mu}_{a,1:t}^{(k-1)}\right) + (t - s) \, \kl\left(\tilde{\mu}_{a,s+1:t}^{(k-1)},\tilde{\mu}_{a,1:t}^{(k-1)}\right) > \beta(t, \delta) \;|\; \cC^{(k-1)}\right) \\
    & \leq \sum_{a=1}^A\bP\left(\exists s < t : s \, \kl(\hat{\mu}_{1:s},\mu_a^{(k-1)}) + (t - s) \, \kl(\hat{\mu}_{s+1:t}, \mu_{a}^{(k-1)}) > \beta(t, \delta)\right),
\end{align*}
with $\hat\mu_{s:s'} = \sum_{r=s}^{s'} Z_{i,r}$ where $Z_{i,r}$ is an \iid{} sequence with mean $\mu_a^{(k-1)}$.
Indeed, conditionally to $\cC^{(k-1)}$, the $n_a(\tau^{(k)})$ successive observations of arm $a$ arm starting from $\hat \tau^{(k)}$ are \iid{} with mean $\mu_a^{(k-1)}$.
Using Lemma~\ref{lem:ConcFirst}, term $(a)$ is upper bounded by $A\delta$.

\paragraph{Upper bound on the term (b): controlling the delay}
From the definition of $\Delta^{c,(k)}$, there exists an arm $a$ such that $\Delta^{c,(k)} = |\mu_a^{(k)} - \mu_a^{(k-1)}|$. We shall prove that it is unlikely that the changepoint detector associated with this arm $a$ doesn't trigger within the delay $d^{(k)}$.
Controlling the detection delay for arm $a$ under the adaptive sampling scheme of \GLRklUCB{} is tricky and we need to leverage the forced exploration (Proposition~\ref{prop:EnoughSamples}) to be sure we have enough samples to ensure detection: the effect is that delays will be scaled by the exploration parameter $\alpha_k$ of the current episode. 

\paragraph{First step: upper bound}
Assume that $\cC^{(k-1)}$ holds. It follows from Proposition~\ref{prop:EnoughSamples} that there exists $\overline{t} \in \{\tau^{(k)}, \dots, \tau^{(k)} + d^{(k)} \}$ such that
$n_a(\overline{t}) - n_a(\tau^{(k)}) = \overline{r}$ where $\overline{r} = \lfloor \frac{\alpha_k}{A} d^{(k)}\rfloor$. This is because the mapping $t\mapsto n_a(t) - n_a(\tau^{(k)})$ is non-decreasing, is $0$ at $t=\tau^{(k)}$ and its value at $\tau^{(k)}+d^{(k)}$ is larger than $\overline{r}$ as $k_t \geq k$.  Using that \[\left(\hat\tau^{(k)} \geq \tau^{(k)} + d^{(k)}\right)\cap \cC^{(k-1)} \subseteq \left(\hat\tau^{(k)} \geq \overline{t}\right)\cap \cC^{(k-1)}\]
further implies that $ (b)=\bP\left(\hat\tau^{(k)} \geq \tau^{(k)} + d^{(k)} | \cC^{(k-1)}\right)$ is upper bounded as follows:
\begin{small}
\[
  (b) \leq \bP\left(\left.n_a(\tau^{(k)}) \, \kl\left(\tilde{\mu}^{k-1}_{i,n_a(\tau^{(k)})},\tilde{\mu}^{k-1}_{i,n_a(\overline{t})}\right)
    + \overline{r} \, \kl\left(\tilde{\mu}^{k-1}_{i,n_a(\tau^{(k)}) : n_a(\overline{t})},\tilde{\mu}^{k-1}_{i,n_a(\overline{t})}\right) \leq \beta(n_a(\tau^{(k)}) + \overline{r},\delta)\right| \cC^{(k-1)}\right),
\]\end{small}
\hspace{-0.2cm}where $\tilde{\mu}^{k-1}_{a,s}$ denotes the empirical mean of the $s$ first observation of arm $a$ since the $(k-1)$-th restart $\hat{\tau}^{(k-1)}$ and  $\tilde{\mu}^{k-1}_{a,s:s'}$ the empirical mean that includes observation number $s$ to number $s'$. Conditionally to $\cC^{(k-1)}$, $\tilde{\mu}^{k-1}_{a,n_a(\tau^{(k)})}$ is the empirical mean of $n_a(\tau^{(k)})$ \iid{} replications of mean $\mu_a^{(k-1)}$, whereas $\tilde{\mu}^{k-1}_{a,n_a(\tau^{(k)}) : n_a(\overline{t})}$ is the empirical mean of $\overline{r}$ \iid{} replications of mean $\mu_a^{(k)}$.

\paragraph{Second step: controlling $\bm{n_a(\tau^{(k)})}$}
Thanks to Proposition~\ref{prop:EnoughSamples}, we know that $n_a(\tau^{(k)})$ lies in the interval $
\left[\left\lfloor \frac{\alpha_k}{A}\left(\tau^{(k)}-\hat\tau^{(k-1)}\right)\right\rfloor,\left(\tau^{(k)}-\hat\tau^{(k-1)}\right)\right]$. Conditionally to $\cC^{(k-1)}$, one obtains furthermore using that $d^{(k-1)} \leq (\tau^{(k)} - \tau^{(k-1)})/2$ (which follows from Assumption~\ref{ass:LongPeriodsGlobal}) that
\begin{eqnarray*}
    n_a(\tau^{(k)})& \in& \cI_k \ \text{ where } \ \cI_k : =\left\{ \left\lfloor \frac{\alpha_k}{2A} (\tau^{(k)}-\tau^{(k-1)} )\right\rfloor, \dots, \tau^{(k)}-\tau^{(k-1)} \right\}.
\end{eqnarray*}
Introducing $\hat{\mu}_{a,s}$ (resp. $\hat{\mu}_{b,s}$) the empirical mean of $s$ \iid{} observations with mean ${\mu}_a^{(k-1)}$ (resp. ${\mu}_a^{(k)}$), such that $\hat{\mu}_{a,s}$ and $\hat{\mu}_{b,r}$ are independent, it follows that
\[
    (b)  \leq \bP\left(\exists s \in \cI_k : s \, \kl\left(\hat{\mu}_{a,s},\frac{s\hat{\mu}_{a,s}+\overline{r}\hat{\mu}_{b,\overline{r}}}{s+\overline{r}} \right) +  \overline{r} \, \kl\left(\hat{\mu}_{b,\overline{r}},\frac{s\hat{\mu}_{a,s}+\overline{r}\hat{\mu}_{b,\overline{r}}}{s+\overline{r}}  \right)  \leq \beta(s+\overline{r},\delta)\right),
\]
where we have also used that $\tilde{\mu}^{k-1}_{a,n_a(\overline{t})} = \left(n_a(\tau^{(k)})\tilde{\mu}^{k-1}_{a,n_a(\tau^{(k)})} + \overline{r}\tilde{\mu}^{k-1}_{a,n_a(\tau^{(k)}) : n_a(\overline{t})} \right) / (n_a(\tau^{(k)}) + \overline{r})$.

\paragraph{Third step: concluding with concentration inequalities} Using Pinsker's inequality and introducing the gap $\Delta_{a}^{c,(k)} = \mu_a^{(k-1)} - {\mu}_a^{(k)}$ (which is such that $\Delta^{c,(k)} = | \Delta_{a}^{c,(k)} |$), one can write
\begin{align}
    &(b) \leq \bP\left(\exists s \in \cI_k : \frac{2s\overline{r}}{s+\overline{r}}\left(\hat{\mu}_{a,s} - \hat{\mu}_{b,\overline{r}} \right)^2  \leq \beta(s+\overline{r},\delta)\right)\nonumber\\
    & \leq \bP\left(\exists s \in \N : \frac{2sr}{s+r}\left(\hat{\mu}_{a,s}\! - \hat{\mu}_{b,s}\! - \Delta_{a}^{c,(k)}\right)^2  \!\!\geq \beta(s+r,\delta)\right) \label{Term1RHS}\\
    & + \bP\left(\!\exists s \in \cI_k : \frac{2s\overline{r}}{s+\overline{r}}\left(\hat{\mu}_{a,s}\! - \hat{\mu}_{b,\overline{r}}\! - \Delta_{a}^{c,(k)}\right)^2 \!\!\! \leq \beta(s+\overline{r},\delta), \frac{2s\overline{r}}{s+\overline{r}}\left(\hat{\mu}_{a,s}\! - \hat{\mu}_{b,\overline{r}}\right)^2  \leq \beta(s+\overline{r},\delta)\right) \label{Term2RHS}
\end{align}
Using Lemma~\ref{lem:Chernoff2arms} (given above in Appendix~\ref{proof:Chernoff2arms}) and a union bound, the first term \eqref{Term1RHS} is upper bounded by $\delta$ (as $\beta(r+s,\delta) \geq \beta(s,\delta) \geq \ln(3s\sqrt{s}/\delta)$). For the second term \eqref{Term2RHS} we use the observation that
\[\frac{2s\overline{r}}{s+\overline{r}}\left(\hat{\mu}_{a,s} - \hat{\mu}_{b,\overline{r}} - \Delta_a^{c,(k)}\right)^2  \leq \beta(s+\overline{r},\delta) \ \ \Rightarrow \ \ |\hat{\mu}_{a,s} - \hat{\mu}_{b,\overline{r}}| \geq |\Delta_a^{c,(k)}| - \sqrt{\frac{s+\overline{r}}{2\overline{r}s}\beta(s+\overline{r},\delta)}\]
and, using that $\Delta^{c,(k)} = | \Delta_a^{c,(k)}|$, one obtains
\[(b) \leq \delta + \bP\left(\exists s \in \cI_k : \Delta^{c,(k)} \leq 2\sqrt{\frac{s+\overline{r}}{2s\overline{r}}\beta(s+\overline{r},\delta)}\right).\]
Let $s_{\min} = \left\lfloor \frac{\alpha_k}{A} (\tau^{(k)}-\tau^{(k-1)})/2\right\rfloor$. Using that the mappings $s \mapsto (s+\overline{r})/s\overline{r}$ and $s \mapsto \beta(s + \overline{r},\delta)$ are respectively decreasing and increasing in $s$, one can further write 
\begin{eqnarray}
 (b) & \leq & \delta + \bP\left(\exists s \in \cI_k : \left(\Delta^{c,(k)}\right)^2 \leq 2\frac{s_{\min}+\overline{r}}{s_{\min}\overline{r}}\beta(\tau^{(k)} - \tau^{(k-1)}+\overline{r},\delta)\right) \nonumber\\
 & \leq & \delta + \bP\left(\exists s \in \cI_k : \left(\Delta^{c,(k)}\right)^2 \leq \frac{4}{\overline{r}}\beta\left(\frac{3}{2}(\tau^{(k)} - \tau^{(k-1)}),\delta\right)\right),\label{FromHereGlobal}
\end{eqnarray}
where in the last step we use that by Assumption~\ref{ass:LongPeriodsGlobal}, it holds that $\overline{r} \leq s_{\min} \leq (\tau^{(k)} - \tau^{(k-1)})/2$. To conclude the proof, it remains to observe that by definition of the delay $d^{(k)}$, 
\[\overline{r} = \left\lfloor \frac{\alpha_k}{A}d^{(k)}\right\rfloor >\frac{4}{\left(\Delta^{c,(k)}\right)^2} \beta\left(\frac{3}{2}(\tau^{(k)} - \tau^{(k-1)}),\delta\right) \]
hence the probability in the right hand side of \eqref{FromHereGlobal} is equal to zero, which yields $(b)\leq \delta$. 

\section{\GLRklUCB{} with Local Restarts}\label{app:Local}

Rather than featuring the number of \emph{breakpoints} $\Upsilon_T$, our analysis for local restarts features the number of \emph{changepoints} $C_T$ defined below. We first define the number of changepoints on arm $a$ as \[\NCi:=\sum_{t=1}^{T-1} \ind\left(\mu_t(a) \neq \mu_{t+1}(a)\right).\] Clearly,  $\NCi \leq \Upsilon_T$, but there can be an arbitrary difference between $\NCi$ and $\Upsilon_T$ for some arms. We denote by $C_T := \sum_{a=1}^A \NCi$ the total number of changepoints on the arms. Observe $C_T$ can take all the  values in $\{\Upsilon_T, \dots, A\Upsilon_T\}$.

We further denote by $\tau_a^{(\ell)}$ the position of the $\ell$-th changepoint \emph{for arm $a$}: \[\tau_a^{(\ell)} = \inf \{ t > \tau_a^{(\ell - 1)} : \mu_a(t) \neq \mu_a(t+1)\},\]
with the convention $\tau_a^{(0)}=0$, and let $\overline{\mu}_a^{(\ell)}$ be the $\ell$-th value for the mean of arm $a$, such that $\forall t \in [\tau_a^{(\ell-1)}+1, \tau_a^{(\ell)}], \ \ \mu_a(t) = \overline{\mu}_a^{(\ell)}$. We also introduce the gap of the $\ell$-th changepoint on arm $a$ which is  $\Delta_a^{c,(\ell)} = \overline{\mu}_a^{\ell} - \overline{\mu}_a^{\ell-1} > 0$.

Assumption~\ref{ass:LongPeriods} requires that any two consecutive changepoints \emph{on a given arm} are sufficiently spaced (relatively to the magnitude of those two changepoints). Under that assumption, Theorem~\ref{thm:mainRegretBound} provides a counterpart to Theorem~\ref{thm:mainRegretBoundGlobal} for \GLRklUCB{} based on Local Restart.

\begin{assumption}\label{ass:LongPeriods}
    Define the delay $d_a^{(\ell)}  =
        d_a^{(\ell)}(\alpha,\delta) = \left\lceil \frac{4A}{\alpha_\ell\left(\Delta_a^{c,(\ell)}\right)^2}\beta\left(\tfrac{3}{2}(\tau_a^{(\ell)} - \tau_a^{(\ell-1)}),\delta\right) + \frac{A}{\alpha_\ell}\right\rceil
    $, we assume that for all arm $a$ and all $\ell \in \{1,\dots,\NCi\}$, $\tau_a^{(\ell)} - \tau_a^{(\ell-1)} \geq 2\max (d_a^{(\ell)},d_a^{(\ell-1)})$.
\end{assumption}

\begin{theorem}
    \label{thm:mainRegretBound}For $\alpha=(\alpha_1,\alpha_2,\dots)$ and $\delta \in (0,1)$ for which Assumption~\ref{ass:LongPeriods} is satisfied, the regret of \GLRklUCB{} with parameters $\alpha$ and $\delta$ based on \textbf{Local} Restart satisfies
    \begin{align*}R_T & \leq (A+2C_T)\delta T + \alpha_{C_T+1} T \\ 
    &+ 2\sum_{a=1}^A\sum_{\ell=1}^{\NCi} \frac{4A}{\alpha_\ell \left(\Delta_a^{c,(\ell)}\right)^2}\beta\left(\tfrac{3}{2}(\tau_a^{(\ell)} - \tau_a^{(\ell-1)}),\delta\right) + \sum_{a=1}^A\sum_{\ell=0}^{\NCi} \frac{\ln(\tau_a^{(\ell+1)} - \tau_a^\ell)}{\kl\left(\overline{\mu}_a^{(\ell)},{\mu}_{a}^{(\ell),*}\right)} +\bigO{\sqrt{\ln(T)}},\end{align*}
    where
    ${\mu}_{a}^{(\ell),*} = \inf \left\{ \mu_{a_t^\star}(t) : \mu_{a_t^\star}(t) \neq \overline{\mu}_a^{(\ell)}, t \in [\tau_a^{(\ell)}+1, \tau_a^{(\ell+1)}]\right\}$, with $\inf \emptyset = 0$.
\end{theorem}

\begin{corollary}\label{cor:Local} Recall $\Delta^{\emph{opt}}$ defined in Section~\ref{sec:Analysis} and let $\Delta$ be the minimum value of $\Delta_a^{c,(\ell)}$ for all $a$ and $\ell \in \{1,\dots,\NCi\}$. For problem instances satisfying the corresponding Assumption~\ref{ass:LongPeriods}, for any $\gamma > 1/2$,
    \begin{enumerate}
        \item Choosing $\alpha_k = \sqrt{\frac{k A\ln(T)}{T}}$ and $\delta = \frac{1}{T^\gamma}$ yields {\small
       \[R_T = \bigO{(1+\gamma)\left(\sum_{a=1}^{A} \sqrt{\NCi}\right)\frac{\sqrt{AT\ln(T)}}{\Delta^2}  + \frac{C_T\ln(T)}{\left(\Delta^{\emph{opt}}\right)^2}}.\]}
        \item If $C_T$ is known, choosing $\alpha_k = \sqrt{\frac{C_T A \ln(T)}{T}}$ and $\delta = \frac{1}{T^\gamma}$ yields {\small\[R_T = \bigO{(1+\gamma)\frac{\sqrt{\EffChange AT\ln(T)}}{\Delta^2} + \frac{\EffChange\ln(T)}{\left(\Delta^{\emph{opt}}\right)^2} }.\]}
    \end{enumerate}
\end{corollary}

Corollary~\ref{cor:Local} specifies possible choices for the exploration sequence $\alpha$ and the parameter $\delta$ that yield $\sqrt{T}$ regret. If the number of changepoint $C_T$ is known, one can achieve $\bigO{\sqrt{C_T A T \ln(T)}/\Delta^2}$, where $\Delta$ is the minimal magnitude of \emph{any} changepoint (which can be smaller than $\Delta^{\text{change}}$ defined in Section~\ref{sec:Analysis}). Without the knowledge of $C_T$, one achieves a slightly worse regret as the $\sqrt{C_T}$ factor is replaced by the larger  $\sum_{a=1}^{A} \sqrt{\NCi}$. As $C_T$ can moreover take any value in $\{\Upsilon_T,\dots,A\Upsilon_T\}$, the guarantees obtained for \GLRklUCB{} with Local Restart are essentially worse than those obtained for Global Restart.  

For particular instances such that $\Upsilon_T = \EffChange$, \ie, at each breakpoint only one arm changes (like in Problem $1$ in Section~\ref{sec:NumericalExperiments}), one obtains however competitive results for \GLRklUCB{} with Local Restart. Indeed, in that case $\Delta = \Delta^{\text{change}}$ and when $\Upsilon_T$ is known the two versions of \GLRklUCB{} achieve the same $\mathcal{O}(\sqrt{\Upsilon_T A T \ln(T)}/(\Delta^{\text{change}})^2)$ regret. Yet when $\Upsilon_T$ is unknown, \GLRklUCB{} with Local restart is only guaranteed to have a regret of $\mathcal{O}(A\sqrt{\Upsilon_T T \ln(T)}/(\Delta^{\text{change}})^2)$ when each arm has the same number of changepoints, which is sub-optimal by a factor $\sqrt{A}$. Still, observe that these similar (or slightly worse) regret guarantees under Local Restart hold for a wider variety of problems as Assumption~\ref{ass:LongPeriods} is less stringent than Assumption~\ref{ass:LongPeriodsGlobal}.

We highlight that the results of Theorem~\ref{thm:mainRegretBoundGlobal} and Theorem~\ref{thm:mainRegretBound} provide only \emph{upper bounds} on the regret, which can be viewed as a sanity-check for using both variants of \GLRklUCB{}. The fact that our current results are worse for Local Restart is not in contradiction with our observation in Section~\ref{sec:NumericalExperiments} that the empirical performance of the two algorithms is very close.

\subsection{Proof of Corollary~\ref{cor:Local}} 
Choosing $\delta =T^{-\gamma}$ and $\alpha_k = \sqrt{{k\ln(T)A}/{T}}$, Theorem~\ref{thm:mainRegretBoundGlobal} upper bound the regret of \GLRklUCB{} by {\footnotesize
\begin{align*}&(A + 2C_T) T^{1-\gamma} + \sqrt{(C_T+1)A T\ln(T)} + \sum_{a=1}^{A}  \sum_{\ell = 1}^{\NCi}\frac{4A}{\alpha_{\ell}(\Delta^{(\ell)}_a)^2} \beta(\tfrac{3}{2}T,T^{-\gamma}) + \sum_{a=1}^{A} \sum_{\ell=1}^{\NCi} \frac{\ln(T)}{\kl\left(\bar\mu_a^{(\ell)},\mu_{a}^{(\ell),*}\right)} + \bigO{\sqrt{\ln(T)}}.
\end{align*}}
\hspace{-0.2cm}For $\gamma > 1/2$, the leading term in this expression is 
\[\sqrt{(C_T+1)A T\ln(T)} + \sum_{a=1}^{A}  \sum_{\ell = 1}^{\NCi}\frac{4A}{\alpha_{\ell}(\Delta^{(\ell)}_a)^2} \beta(\tfrac{3}{2}T,T^{-\gamma}) + \sum_{a=1}^{A} \sum_{\ell=1}^{\NCi} \frac{\ln(T)}{\kl\left(\bar\mu_a^{(\ell)},\mu_{a}^{(\ell),*}\right)}.\]
Using further that there exists some absolute constant $C'$ such that $\beta(n,\delta) \leq C' \ln(n/\delta)$, one obtains
\begin{equation}R_T = \bigO{\sqrt{(C_T+1)A T\ln(T)} + (1+\gamma)\sum_{a=1}^{A}  \sum_{\ell = 1}^{\NCi}\frac{A}{\alpha_{\ell}\Delta^{(\ell)}_a}\ln(T) + \frac{C_T\ln(T)}{\left(\Delta^{\text{opt}}\right)^2}}\label{tooptim:local}\end{equation}
Finally, the first statement of the corollary follows from the fact that, for all $i$,  
\[\sum_{\ell=1}^{\NCi}\frac{1}{\alpha_\ell} = \sqrt{\frac{T}{A\ln(T)}}\sum_{\ell=1}^{\NCi} \frac{1}{\sqrt{\ell}} \leq \sqrt{\frac{\NCi T}{A\ln(T)}}.\]
Choosing $\alpha_\ell = \sqrt{\frac{AC_T\ln(T)}{T}}$ in \eqref{tooptim:local} yields the second statement.

\subsection{Proof of Theorem~\ref{thm:mainRegretBound}}

We first introduce some notation for the proof. Recall that $\tau_a^{(\ell)}$ denotes the $\ell$-th changepoint for arm $a$. We use the convention $\tau_a^{(\NCi+1)} = T$. We denote by $\hat\tau_a^{(\ell)}$ the $\ell$-th changepoint detected for arm $a$ by \GLRklUCB, leading to a restart for this arm. 

Distinguishing the exploration steps and the steps in which \GLRklUCB{} uses the UCBs to select the next arm to play, one can upper bound the regret as{\small
\begin{align} R_T  \leq  \bE\left[\sum_{t=1}^T \ind{\left(t \left\lceil \frac{A}{\alpha_{k_t}} \right\rceil \in \{1,\dots,A\}\right)} +  \sum_{t=1}^T (\mu_{a_t^\star}(t) - \mu_{A_t}(t))\ind{\left(\UCB_{A_t}(t-1) \geq \UCB_{a_t^\star}(t-1)\right)}\right] \label{twoterms}
\end{align}}
\hspace{-0.2cm}
We now introduce some high-probability event in which all the \emph{changepoints} are detected within a reasonable delay for all arms. With $d_a^{(\ell)}=d_a^{(\ell)}(\alpha,\delta)$ in Assumption~\ref{ass:LongPeriods}, we define
\[
    \cE_T = \cE_T(\alpha,\delta) = \left(\forall a \in \{1, \ldots, A\}, \forall \ell \in \{1, \ldots, \NCi\}, \hat{\tau}^{(\ell)}_a \in \left[\tau_a^{(\ell)} + 1, \tau_a^{(\ell)} + d_a^{(\ell)}\right] , \hat\tau_a^{(\NCi +1)}>T\right).
\]
From Assumption~\ref{ass:LongPeriods}, as the period between two changepoints are long enough, if $\cE_T$ holds, then for all arm $a$ and all changepoint $\ell$, one has $\tau_a^{(\ell)} \leq \hat{\tau}_a^{(\ell)} \leq \tau_a^{(\ell+1)}$ for all $\ell \in \{1,\dots,\NCi\}$. Also, when $\cE_T$ holds, \GLRklUCB{} experiences a total of $C_T$ restarts (on different arms), which each yield an update of the exploration parameter. Letting $\hat \sigma^{(k)}$ be the instant of the $k$-th restart (on any arm) with the convention that $\hat \sigma^{(C_T+1)} = T$, one can write, when $\cE_T$ holds: 
\begin{eqnarray*}\sum_{t=1}^T \ind{\left(t \left\lceil \frac{A}{\alpha_{k_t}} \right\rceil \in \{1,\dots,A\}\right)} &\leq& \sum_{k=0}^{C_T} \sum_{t=\hat{\sigma}^{(k)} +1}^{\hat\sigma^{(k+1)}} \ind\left(t \left\lceil \frac{A}{\alpha_{k+1}} \right\rceil \in \{1,\dots,A\}\right) \\
&\leq& \sum_{k=0}^{C_T} \alpha_{k+1}\left(\hat{\sigma}^{(k+1)} - \hat{\sigma}^{(k)}\right) \leq \alpha_{C_T + 1 }\sum_{k=0}^{C_T} \left(\hat{\sigma}^{(k+1)} - \hat{\sigma}^{(k)}\right) \\
& = & \alpha_{C_T + 1 }  T.
\end{eqnarray*}
On $\cE_T$, the second term in \eqref{twoterms} can be further decomposed as follows, according to whether the upper confidence bound of the current optimal arm is small, leading to 
{\small
\begin{align*} R_T &\leq  T \bP\left(\cE_T^c\right) + \alpha_{C_T + 1 }  T \nonumber + \\ & \ \ \ \underbrace{\bE\left[\ind(\cE_T) \sum_{t=1}^T \ind\left(\UCB_{a^\star_t}(t) \leq \mu_{a^\star_t}(t)\right)\right]}_{(C)} + \ \underbrace{\bE\left[\ind(\cE_T)\sum_{t=1}^T (\mu_{a^\star_t}(t) - \mu_{A_t}(t))\ind\left(\UCB_{A_t}(t) \geq \mu_{a^\star_t}(t)\right)\right]}_{(D)}.
\end{align*}}

As in the previous proof, the conclusion follows from two lemmas. Lemma~\ref{lem:bandit} controls terms $(C)$ and $(D)$ using some elements from the analysis of \klUCB{} while Lemma~\ref{lem:GoodEvent} upper bounds the probability of the ``bad'' event $\cE_T^c$. The proofs of these two results are presented in the next sections. 

\begin{lemma}\label{lem:bandit} It holds that
\begin{eqnarray}
    (C) &\leq& \sum_{a=1}^A \sum_{\ell=0}^{\NCi}d_a^{(\ell)}(\alpha,\delta) + C_T\ln(\ln(T))\;,
    \label{TermAFinal} \\
    (D)  &\leq& \sum_{a=1}^A\sum_{\ell=0}^{\NCi}d_a^{(\ell)}(\alpha,\delta)+\sum_{a=1}^A\sum_{\ell=1}^{\NCi}\left[ \frac{\ln(\tau_a^{(\ell+1)} - \tau_a^{(\ell)})}{\kl(\overline{\mu}_a^{(\ell)},\mu_{a}^{(\ell),*})} + O\left(\sqrt{\ln(T)}\right)\right]\;. \label{TermBFinal}
\end{eqnarray}
\end{lemma}

\begin{lemma}\label{lem:GoodEvent}
    Under Assumption~\ref{ass:LongPeriods}, it holds that $\bP(\cE_T^c) \leq (A+2\EffChange)\delta$.
\end{lemma}

\subsection{Proof of Lemma~\ref{lem:bandit}}

\paragraph{Upper bound on the term (C)}
\begin{align*}
    (C) & \leq \bE\left[\ind(\cE_T) \sum_{t=1}^T \ind\left(n_{a_t^\star}(t) \kl\left(\hat{\mu}_{a_t^\star}(t), \mu_{a_t^\star}(t)\right) \geq f(t - \tau_{a_t^\star}(t))\right)\right] \\
    & \leq \sum_{a=1}^A \bE\left[\ind(\cE_T) \sum_{t=1}^T \ind(a_t^\star =a) \ind\left(n_{a}(t) \kl\left(\hat{\mu}_{a}(t), \mu_{a}(t)\right) \geq f(t - \tau_{a}(t))\right)\right] \\
    & \leq \sum_{a=1}^A \sum_{\ell=0}^{\NCi}\bE\left[\ind(\cE_T) \sum_{t={\tau}_a^{(\ell)}+1}^{\tau_a^{(\ell+1)}} \ind\left(n_{a}(t) \kl\left(\hat{\mu}_{a}(t), \bar\mu_a^{(\ell)}\right) \geq f(t - \hat\tau_{a}(t))\right)\right] \\
    & \leq \sum_{a=1}^A \sum_{\ell=0}^{\NCi}d_a^{(\ell)}(\alpha,\delta) + \sum_{a=1}^A \sum_{\ell=0}^{\NCi}\bE\left[\ind(\cC_a^{(\ell)}) \sum_{t=\hat{\tau}_a^{(\ell)}+1}^{\tau_a^{(\ell+1)}}\ind\left(n_{a}(t) \kl\left(\hat{\mu}_{a}(t), \overline{\mu}_{a}^{(\ell)}\right) \geq f(t - \hat\tau_{a}^{(\ell)})\right)\right],
\end{align*}%
where we introduce the event $\cC_a^{(\ell)}$ that all the changepoints on arm $a$ up to the $\ell$-th have been detected:
\begin{equation}\cC_a^{(\ell)} = \left\{\forall j \leq \ell, \hat\tau_a^{(j)} \in \left[\tau_a^{(j)} + 1, \tau_a^{(j)} + d_a^{(\ell)} \right] \right\}.\label{def:EventCi}\end{equation}
Clearly, $\cE_T \subseteq \cC_a^{(\ell)}$ and $\cC_a^{(\ell)}$ is $\cF_{\hat\tau_a^{(\ell)}}$-measurable. Observe that conditionally to $\cF_{\hat\tau_a^{(\ell)}}$, when $\ind(\cC_a^{(\ell)})$ holds, $\hat{\mu}_{a}(t)$ is the average of samples that have all mean $\bar\mu_a^{(\ell)}$. Thus, introducing $\hat{\mu}_s$ as a sequence of \iid{} random variables with mean $\bar\mu_a^{(\ell)}$, one can write
\begin{align*}
    &\bE\left[\left.\ind(\cC_a^{(\ell)}) \sum_{t=\hat{\tau}_i^{(\ell)}+1}^{\tau_a^{(\ell+1)}} \ind\left(n_{a}(t) \kl\left(\hat{\mu}_{a}(t), \bar\mu_a^{(\ell)}\right) \geq f(t - \hat\tau_{i}^{(\ell)})\right) \right| \cF_{\hat\tau_a^{(\ell)}}\right]\\
    & = \bE\left[\ind(\cC_a^{(\ell)}) \sum_{t=\hat{\tau}_i^{(\ell)}+1}^{\tau_a^{(\ell+1)}} \bE\left[\ind\left(n_{a}(t) \kl\left(\hat{\mu}_{a}(t), \bar\mu_a^{(\ell)}\right) \geq f(t - \hat\tau_{i}^{(\ell)})\right) \;|\; \cF_{\hat\tau_a^{(\ell)}}\right]\right]\\
    & \leq \bE\left[\ind(\cC_a^{(\ell)}) \sum_{t'=1}^{\tau_a^{(\ell+1)} - \hat{\tau}_i^{(\ell)}} \bP\left(\exists s \leq t' : s \times \kl(\hat{\mu}_{s},\bar\mu_a^{(\ell)}) \geq f(t')\right)\right] \leq 2+\sum_{t=3}^T \frac{1}{t\ln(t)} \leq 3+\ln(\ln(T)),
\end{align*}%
where the last but one inequality relies on the concentration inequality given in Lemma 2 of \cite{KLUCBJournal}, and the fact that $f(t) = \ln(t) + 3 \ln(\ln(t))$.

\paragraph{Upper bound on the term (D)}
Recall that $\mu_{a}^{(\ell),*}$ is defined in the statement of Theorem~\ref{thm:mainRegretBound} as the smallest value of $\mu_{a_t^\star}(t)$ when arm $a$ is sub-optimal on the interval $[\tau^{(\ell)}_a+1, \tau^{(\ell +1)}_a]$. We let $\tilde{\mu}_{a,s}^{(\ell)}$ denote the empirical mean of the first $s$ observations of arm $a$ made after time $t=\hat{\tau}_a^{(\ell)}+1$. To upper bound Term (D), we introduce a sum over all arms and rewrite the sum in $t$ as a sum of consecutive intervals $[\tau_a^{(\ell)}+1, \tau_a^{(\ell+1)}]$.
\begin{align*}
    (D) & \leq \sum_{a=1}^A\bE\Big[\ind(\cE_T)\sum_{\ell=0}^{\NCi} \sum_{t=\tau_a^{(\ell)}+1}^{\tau_a^{(\ell+1)}}\left(\mu_{a^\star_t}(t) - \overline{\mu}_{a}^{(\ell)}\right)\ind\left(A_t = a, \UCB_a(t) \geq \mu_{a_t^\star}(t)\right)\Big] \\
    & \leq \sum_{a=1}^A\bE\Big[\ind(\cE_T)\sum_{\ell=0}^{\NCi} \sum_{t=\tau_a^{(\ell)}+1}^{\tau_a^{(\ell+1)}}\left(\mu_{a^\star_t}(t) - \overline{\mu}_{a}^{(\ell)}\right)\ind\left(A_t = a, \UCB_a(t) \geq \mu_{a}^{(\ell),*}\right)\Big] \\
    & \leq \sum_{a=1}^A\sum_{\ell=0}^{\NCi}\bE\Big[\ind(\cE_T)\hat{\tau}_i^{(\ell)} + \ind(\cE_T) \sum_{t=\hat\tau_a^{(\ell)}+1}^{\tau_a^{(\ell+1)}}\overline{\Delta}_a^{(\ell)}\ind\left(A_t = a, \UCB_a(t) \geq \mu_{a}^{(\ell),*}\right)\Big] 
\end{align*}
with $\overline{\Delta}_a^{(\ell)} := \underset{t \in [\tau_a^{(\ell)}+1,\tau_a^{(\ell +1)}]}{\max} \left(\mu_{a^\star_t}(t) - \overline{\mu}_{a}^{(\ell)}\right)$. Introducing a sum over $\ind(n_a(t) = s)$ and swapping the sums yields 
{\small \begin{align*}
    (D) & \leq \sum_{a=1}^A\sum_{\ell=0}^{\NCi}d_a^{(\ell)}(\alpha,\delta)+ \sum_{a=1}^A\sum_{\ell=1}^{\NCi}\overline{\Delta}_a^{(\ell)}\bE\Big[\ind(\cC_a^{(\ell)}) \sum_{s=1}^{n_a(\tau_a^{(\ell+1)})}\ind\left(s \times \kl(\tilde{\mu}_{a,s}^{(\ell)}, \mu_{a}^{(\ell),*}) \leq f(\tau_a^{(\ell+1)} - \tau_a^{(\ell)})\right)\Big].
\end{align*}}
\hspace{-0.2cm} Conditionally to $\cF_{\hat\tau_a^{(\ell)}}$, when $\cC_a^{(\ell)}$ holds, for $s \in \{1, \ldots, n_a(\tau_a^{(\ell+1)})\}$, $\tilde{\mu}_{a,s}^{(\ell)}$ is the empirical mean from \iid{} observations of mean $\overline{\mu}_a^{(\ell)}$.
Therefore, introducing $\hat{\mu}_s$ as a sequence of \iid{} random variables with mean $\overline{\mu}_a^{(\ell)}$, it follows from the law of total expectation that
\begin{align*}
    (D) & \leq \sum_{a=1}^A\sum_{\ell=0}^{\NCi} d_a^{(\ell)}(\alpha,\delta)
    + \sum_{a=1}^A\sum_{\ell=1}^{\NCi} \overline{\Delta}_i^{(\ell)} \times \!\!\!\!\sum_{s=1}^{\tau_a^{(\ell+1)}-\tau_a^{(\ell)}}\!\!\! \bP\left(s \times \kl(\hat{\mu}_{s}, \mu_{a}^{(\ell),*}) \leq f(\tau_a^{(\ell+1)} - \tau_a^{(\ell)})\right).
\end{align*}%
If $\overline{\Delta}_a^{(\ell)} \neq 0$, then $\mu_{a}^{(\ell),*} > \bar\mu_a^{(\ell)}$ and we can use the same analysis as in the proof of Fact~2 in Appendix~A.2 of \cite{KLUCBJournal} to show that
\[\sum_{s=1}^{\tau_a^{(\ell+1)}-\tau_a^{(\ell)}}\!\!\! \bP\left(s \times \kl(\hat{\mu}_{s}, \mu_{a}^{(\ell),*}) \leq f(\tau_a^{(\ell+1)} - \tau_a^{(\ell)})\right)\leq \frac{\ln(\tau_a^{(\ell+1)} - \tau_a^{(\ell)})}{\kl\left(\overline{\mu}_a^{(\ell)},\mu_{a}^{(\ell),*}\right)} + O\left(\sqrt{\ln(T)}\right)\;.\]
If $\overline{\Delta}_a^{(\ell)} = 0$, then $\mu_{a}^{(\ell),*}=0$ and $\kl\left(\overline{\mu}_a^{(\ell)},\mu_{a}^{(\ell),*}\right) = +\infty$, thus it also trivially holds that 
\[\overline{\Delta}_a^{(\ell)}\sum_{s=1}^{\tau_a^{(\ell+1)}-\tau_a^{(\ell)}}\!\!\! \bP\left(s \times \kl(\hat{\mu}_{s}, \mu_{a}^{(\ell),*}) \leq f(\tau_a^{(\ell+1)} - \tau_a^{(\ell)})\right)\leq \frac{\ln(\tau_a^{(\ell+1)} - \tau_a^{(\ell)})}{\kl\left(\overline{\mu}_a^{(\ell)},\mu_{a}^{(\ell),*}\right)} + O\left(\sqrt{\ln(T)}\right)\;\]
and the proof of \eqref{TermBFinal} is complete.

\subsection{Proof of Lemma~\ref{lem:GoodEvent}}\label{proof:GoodEvent}

With the event $\cC_a^{(\ell)}$ defined in \eqref{def:EventCi} and the convention $\tau_a^{(\NCi +1)} = T$, a simple union bound yields
\begin{align*}
    \bP(\cE_T^c) & \leq \sum\limits_{a=1}^A\sum\limits_{\ell=1}^{\NCi+1} \underbrace{\bP\left(\left.\hat\tau_a^{(\ell)} \leq \tau_a^{(\ell)} \;\right|\; \cC_a^{(\ell-1)}\right)}_{(a)}  + \sum\limits_{i=1}^A\sum\limits_{\ell=1}^{\NCi} \underbrace{\bP\left(\left.\hat\tau_a^{(\ell)} \geq \tau_a^{(\ell)} + d_a^{(\ell)} \;\right|\; \cC_a^{(\ell-1)}\right)}_{(b)}.
\end{align*}
The final result follows by proving that the terms $(a)$ and $(b)$ are both upper bounded by $\delta$.

\paragraph{Upper bound on $(a)$: controlling the false alarms} $\hat\tau_a^{(\ell)} \leq \tau_a^{(\ell)}$ implies that there is a false alarm for the detection of the $\ell$-th change on arm $a$, which is not likely: 
\begin{align*}
    (a) & \leq \bP\left(\exists s < t \leq n_a(\tau_a^{(\ell)}) : s \times \kl\left(\tilde{\mu}_{a,1:s}^{(\ell-1)},\tilde{\mu}_{a,1:t}^{(\ell-1)}\right) + (t - s) \times \kl\left(\tilde{\mu}_{i,s+1:t}^{(\ell-1)},\tilde{\mu}_{a,1:t}^{(\ell-1)}\right) > \beta(t, \delta) \;|\; \cC_a^{(\ell-1)}\right) \\
    & \leq \bP\left(\exists s < t : s \times \kl(\hat{\mu}_{1:s},\mu_a^{(\ell-1)}) + (t - s) \times \kl(\hat{\mu}_{s+1:t}, \mu_{a}^{(\ell-1)}) > \beta(t, \delta)\right),
\end{align*}
with $\hat\mu_{s:s'} = \sum_{r=s}^{s'} Z_{i,r}$ where $Z_{i,r}$ is an \iid{} sequence with mean $\mu_a^{(\ell-1)}$.
Indeed, conditionally to $\cC_a^{(\ell-1)}$, the $n_a(\tau_a^{(\ell)})$ successive observations of arm $a$ arm starting from $\hat \tau_a^{(\ell)}$ are \iid{} with mean $\mu_a^{(\ell-1)}$.
Using Lemma~\ref{lem:ConcFirst}, term $(a)$ is upper bounded by $\delta$.

\paragraph{Upper bound on term $(b)$: controlling the delay}
Assume that $\cC_a^{(\ell-1)}$ holds. For $t > \hat\tau_a^{(\ell-1)}$, we now that at least the first $\ell-1$ changepoints on arm $a$ have been detected, hence the current number of episodes $k_t$ is larger than $\ell$. It follows from Proposition~\ref{prop:EnoughSamples} that there exists $\overline{t} \in \left\{\tau_a^{(\ell)}, \dots, \tau_a^{(\ell)} + d_a^{(\ell)} \right\}$ such that
$n_a(\overline{t}) - n_a(\tau_a^{(\ell)}) = \overline{r}$ where $\overline{r} = \lfloor \frac{\alpha_\ell}{A} d_a^{(\ell)}\rfloor$. This is because the mapping $t\mapsto n_a(t) - n_a(\tau_a^{(\ell)})$ is non-decreasing, is $0$ at $t=\tau_a^{(\ell)}$ and its value at $\tau_a^{(\ell)}+d_a^{(\ell)}$ is larger than $\overline{r}$ by Proposition~\ref{prop:EnoughSamples} as $\alpha_{k_t} \geq \alpha_{\ell}$.
Using that \[(\hat\tau_a^{(\ell)} \geq \tau_a^{(\ell)} + d_a^{(\ell)})\cap \cC_a^{(\ell-1)} \subseteq (\hat\tau_a^{(\ell)} \geq \overline{t})\cap \cC_a^{(\ell-1)},\]
the probability $(b)=\bP(\hat\tau_a^{(\ell)} \geq \tau_a^{(\ell)} + d_a^{(\ell)} | \cC_a^{(\ell-1)})$ is further upper bounded as follows: {\small
\[
 (b) \leq \bP\left(\left. n_a(\tau_a^{(\ell)}) \, \kl\left(\tilde{\mu}^{\ell-1}_{i,n_a(\tau_a^{(\ell)})},\tilde{\mu}^{\ell-1}_{i,n_a(\overline{t})}\right)
    +  \overline{r} \, \kl\left(\tilde{\mu}^{\ell-1}_{i,n_a(\tau_a^{(\ell)}) : n_a(\overline{t})},\tilde{\mu}^{\ell-1}_{i,n_a(\overline{t})}\right) \leq \beta(n_a(\tau_a^{(\ell)}) + \overline{r},\delta) \right| \cC_a^{(\ell-1)} \right),
\]} \hspace{-0.2cm}
where $\tilde{\mu}^{\ell-1}_{i,s}$ denotes the empirical mean of the $s$ first observation of arm $a$ since the $(\ell-1)$-th restart $\hat{\tau}_i^{(\ell-1)}$ and  $\tilde{\mu}^{\ell-1}_{i,s:s'}$ the empirical mean that includes observation number $s$ to number $s'$. Conditionally to $\cC^{(\ell-1)}_i$, $\tilde{\mu}^{\ell-1}_{i,n_a(\tau_a^{(\ell)})}$ is the empirical mean of $n_a(\tau_a^{(\ell)})$ \iid{} replications of mean $\bar\mu_i^{\ell-1}$, whereas $\tilde{\mu}^{\ell-1}_{i,n_a(\tau_a^{(\ell)}) : n_a(\overline{t})}$ is the empirical mean of $\overline{r}$ \iid{} replications of mean $\bar\mu_i^{\ell}$.

Then, conditionally to $\cC_a^{(\ell-1)}$,  $n_a(\tau_a^{(\ell)}) \in
\left\{\left\lfloor \frac{\alpha_\ell}{A}\left(\tau_a^{(\ell)}-\hat\tau_a^{(\ell-1)}\right)\right\rfloor, \dots, \tau_a^{(\ell)}-\hat\tau_a^{(\ell-1)} \right\}$ due to Proposition~\ref{prop:EnoughSamples} and to the fact that $d_a^{(\ell-1)} \leq (\tau_a^{(\ell)} - \tau_a^{(\ell-1)})/2$ by Assumption~\ref{ass:LongPeriods}. Hence
\begin{eqnarray*}
    n_a(\tau_a^{(\ell)})& \in &  \left\{ \left\lfloor \frac{\alpha_\ell}{A}\left(\tau_a^{(\ell)}-\tau_a^{(\ell-1)} - d_a^{(\ell-1)}\right)\right\rfloor, \dots, \left(\tau_a^{(\ell)}-\tau_a^{(\ell-1)}\right) \right\} \\
    n_a(\tau_a^{(\ell)})& \in& \left\{ \left\lfloor \frac{\alpha_\ell}{2A}\left(\tau_a^{(\ell)}-\tau_a^{(\ell-1)}\right)\right\rfloor, \dots, \left(\tau_a^{(\ell)}-\tau_a^{(\ell-1)}\right) \right\} := \cI_\ell.
\end{eqnarray*}

Introducing $\hat{\mu}_{a,s}$ (resp. $\hat{\mu}_{b,s}$) the empirical mean of $s$ \iid{} observations with mean $\bar{\mu}_a^{(\ell-1)}$ (resp. $\bar{\mu}_a^{(\ell)}$), such that $\hat{\mu}_{a,s}$ and $\hat{\mu}_{b,r}$ are independent, it follows that
\[
    (b)  \leq \bP\left(\exists s \in \cI_{\ell} : s \, \kl\left(\hat{\mu}_{a,s},\frac{s\hat{\mu}_{a,s}+\overline{r}\hat{\mu}_{b,\overline{r}}}{s+\overline{r}} \right) +  \overline{r} \, \kl\left(\hat{\mu}_{b,\overline{r}},\frac{s\hat{\mu}_{a,s}+\overline{r}\hat{\mu}_{b,\overline{r}}}{s+\overline{r}}  \right)  \leq \beta(s+\overline{r},\delta)\right),
\]
where we have also used that $\tilde{\mu}^{\ell-1}_{a,n_a(\overline{t})} = \frac{n_a(\tau_a^{(\ell)})\tilde{\mu}^{\ell-1}_{a,n_a(\tau_a^{(\ell)})} + \overline{r}\tilde{\mu}^{\ell-1}_{a,n_a(\tau_a^{(\ell)}) : n_a(\overline{t})}}{n_a(\tau_a^{(\ell)}) + \overline{r}}$.

Using Pinsker's inequality and the expression of the gap $\Delta_a^{c,(\ell)} = \bar\mu_a^{(\ell-1)} - \bar{\mu}_a^{(\ell)}$, one can write
\begin{align}
    (b) & \leq \bP\left(\exists s \in \cI_{\ell} : \frac{2s\overline{r}}{s+\overline{r}}\left(\hat{\mu}_{a,s} - \hat{\mu}_{b,\overline{r}} \right)^2  \leq \beta(s+\overline{r},\delta)\right)\nonumber\\
    & \leq \bP\left(\exists s \in \N : \frac{2sr}{s+r}\left(\hat{\mu}_{a,s} - \hat{\mu}_{b,s} - \Delta_a^{c,(\ell)}\right)^2  \geq \beta(s+r,\delta)\right) \nonumber\\
    & + \bP\left(\exists s \in \cI_{\ell} : \frac{2s\overline{r}}{s+\overline{r}}\left(\hat{\mu}_{a,s} - \hat{\mu}_{b,\overline{r}} - \Delta_a^{c,(\ell)}\right)^2  \leq \beta(s+\overline{r},\delta), \frac{2s\overline{r}}{s+\overline{r}}\left(\hat{\mu}_{a,s} - \hat{\mu}_{b,\overline{r}}\right)^2  \leq \beta(s+\overline{r},\delta)\right) \nonumber
\end{align}

Using Lemma~\ref{lem:Chernoff2arms} stated in Appendix~\ref{proof:Chernoff2arms} and a union bound, the first term in the right hand side is upper bounded by $\delta$ (as $\beta(r+s,\delta) \geq \beta(s,\delta) \geq \ln(3s\sqrt{s}/\delta)$). For the second term, we use the observation
\[\frac{2s\overline{r}}{s+\overline{r}}\left(\hat{\mu}_{a,s} - \hat{\mu}_{b,\overline{r}} - \Delta_a^{c,(\ell)}\right)^2  \leq \beta(s+\overline{r},\delta) \ \ \Rightarrow \ \ |\hat{\mu}_{a,s} - \hat{\mu}_{b,\overline{r}}| \geq |\Delta_a^{c,(\ell)}| - \sqrt{\frac{s+\overline{r}}{2\overline{r}s}\beta(s+\overline{r},\delta)}\]
and finally get
\[(b) \leq \delta + \bP\left(\exists s \in \cI_\ell : |\Delta_a^{c,(\ell)}| \leq 2\sqrt{\frac{s+\overline{r}}{2s\overline{r}}\beta(s+\overline{r},\delta)}\right).\]
Let $s_{\min} = \left\lfloor \frac{\alpha_\ell}{A} (\tau_a^{(\ell)}-\tau_a^{(\ell-1)})/2\right\rfloor$. Using that the mappings $s \mapsto (s+\overline{r})/s\overline{r}$ and $s \mapsto \beta(s + \overline{r},\delta)$ are respectively decreasing and increasing in $s$, one can further write 
\begin{eqnarray}
 (b) & \leq & \delta + \bP\left(\exists s \in \cI_\ell : \left(\Delta_a ^{(\ell)}\right)^2 \leq 2\frac{s_{\min}+\overline{r}}{s_{\min}\overline{r}}\beta(\tau_a^{(\ell)} - \tau_a^{(\ell-1)}+\overline{r},\delta)\right) \nonumber\\
 & \leq & \delta + \bP\left(\exists s \in \cI_\ell : \left(\Delta_a^{(\ell)}\right)^2 \leq \frac{4}{\overline{r}}\beta\left(\frac{3}{2}(\tau_a^{(\ell)} - \tau_a^{(\ell-1)}),\delta\right)\right),\label{FromHereLocal}
\end{eqnarray}
where in the last step we use that by Assumption~\ref{ass:LongPeriods}, it holds that $\overline{r} \leq s_{\min} \leq (\tau_a^{(\ell)} - \tau_a^{(\ell-1)})/2$. To conclude the proof, it remains to observe that by definition of the delay $d_a^{(\ell)}$, 
\[\overline{r} = \left\lfloor \frac{\alpha_\ell}{A}d_a^{(\ell)}\right\rfloor >\frac{4}{\left(\Delta_a^{c,(\ell)}\right)^2} \beta\left(\frac{3}{2}(\tau_a^{(\ell)} - \tau_a^{(\ell-1)}),\delta\right) \]
hence the probability in the right hand side of \eqref{FromHereLocal} is equal to zero, which yields $(b)\leq \delta$. 

\end{document}